\documentclass[11pt]{article}

\usepackage[preprint]{acl}

\usepackage{iftex}
\ifPDFTeX
  \usepackage{times}
  \usepackage[T1]{fontenc}
  \usepackage[utf8]{inputenc}
  \newcommand{\ko}[1]{\textit{[Korean text; build with xelatex to typeset it]}}
\else
  \usepackage{fontspec}
  \defaultfontfeatures{Extension = .otf}
  \newfontfamily\koreanfont{NotoSansKR-subset}[
    Path = fonts/,
    UprightFont = *-Regular, BoldFont = *-Bold,
    Scale = MatchLowercase]
  \newcommand{\ko}[1]{{\koreanfont #1}}
\fi

\usepackage{latexsym}
\usepackage{microtype}
\usepackage{booktabs}
\usepackage{graphicx}
\usepackage{xcolor}
\usepackage{amsmath}
\usepackage{url}
\usepackage{orcidlink}

\makeatletter
\newif\ifresultspending
\resultspendingfalse
\newcommand{\RESULT}[1]{%
  \immediate\write\@auxout{\string\global\string\resultspendingtrue}%
  \textcolor{red}{\textbf{[RESULT: #1]}}%
}
\makeatother

\makeatletter
\newcommand{\pendingflag}{\immediate\write\@auxout{\string\global\string\resultspendingtrue}}
\newcommand{\artifactstub}[1]{%
  \begin{center}
    \fbox{\parbox{0.9\columnwidth}{\centering\small\color{red}%
      \texttt{\detokenize{#1}} not generated yet.\\ Run \texttt{make analyze}.}}
  \end{center}%
}
\makeatother

\newcommand{\resulttable}[1]{%
  \IfFileExists{tables/#1.tex}{\input{tables/#1}}{\pendingflag\artifactstub{tables/#1.tex}}%
}

\newcommand{\resultnum}[1]{\pendingflag\textcolor{red}{\textbf{[NUM: \detokenize{#1}]}}}
\newcommand{\resultfigure}[2]{%
  \IfFileExists{figures/#1}{\includegraphics[width=#2]{figures/#1}}%
                           {\pendingflag\artifactstub{figures/#1}}%
}

\newcommand{\draftbanner}{%
  \ifresultspending
    \begin{center}
      \colorbox{red!12}{%
        \parbox{0.92\columnwidth}{\centering\bfseries\color{red}%
          DRAFT: RESULTS PENDING\\[2pt]
          \normalfont\small
          Unfilled \texttt{[RESULT: \ldots]} placeholders remain in the
          abstract, or a generated table or figure is missing.
          This manuscript is not submittable.}%
      }
    \end{center}
    \vspace{4pt}
  \fi
}

\title{Repair, Not Improvement: Decomposing Constrained\\
       Decoding in Tool-Call Abstention}

\hypersetup{%
  pdftitle={Repair, Not Improvement: Decomposing Constrained Decoding in Tool-Call Abstention},
  pdfauthor={Janghoon Lee},
}

\author{Janghoon Lee\,\orcidlink{0009-0002-8108-5407} \\
        Redrob \\
        \texttt{janghoon@redrob.io} \\
        {\small ORCID 0009-0002-8108-5407}}

\begin{document}

\IfFileExists{tables/table_macros.tex}{
\makeatletter
\renewcommand{\resultnum}[1]{%
  \@ifundefined{resultnum@#1}%
    {\PackageError{results}{no such result key: #1}{}}%
    {\csname resultnum@#1\endcsname}%
}
\@namedef{resultnum@dmask/abstain/min}{$-24.0$}
\@namedef{resultnum@dmask/abstain/max}{$+19.5$}
\@namedef{resultnum@dmask/abstain/min-cell}{Qwen3-1.7B, English}
\@namedef{resultnum@dmask/abstain/max-cell}{Qwen3-0.6B, Korean}
\@namedef{resultnum@dmask/tool/min}{$+5.9$}
\@namedef{resultnum@dmask/tool/max}{$+32.5$}
\@namedef{resultnum@dmask/tool/min-cell}{Qwen3-4B, English}
\@namedef{resultnum@dmask/tool/max-cell}{Qwen3-0.6B, Korean}
\@namedef{resultnum@paired/min}{$-22.7$}
\@namedef{resultnum@paired/max}{$+2.7$}
\@namedef{resultnum@paired/min-cell}{Qwen3-0.6B}
\@namedef{resultnum@paired/max-cell}{Qwen3-1.7B}
\@namedef{resultnum@dmask/abstain/n-positive}{two}
\@namedef{resultnum@dmask/abstain/n-cells}{six}
\@namedef{resultnum@repair/n-ft}{698}
\@namedef{resultnum@repair/n-ft-unreadable}{545}
\@namedef{resultnum@repair/n-ft-judgement}{0}
\@namedef{resultnum@repair/cell}{Qwen3-0.6B KO}
\@namedef{resultnum@repair/cell-n-ft}{330}
\@namedef{resultnum@repair/cell-n-ft-unreadable}{325}
\@namedef{resultnum@decomposition/gap}{$+0.083$ $[-0.186, +0.300]$}
\@namedef{resultnum@position/abstain-shift-max}{$+3.0$}
\@namedef{resultnum@position/abstain-shift-mean}{$+0.5$}
\@namedef{resultnum@position/tool-not-above-free}{five}
\@namedef{resultnum@position/gold-not-first-n}{0}
\@namedef{resultnum@position/gold-n}{200}
\@namedef{resultnum@census/total}{145{,}969}
\@namedef{resultnum@census/confirmatory}{36{,}000}
\@namedef{resultnum@census/regression}{52{,}000}
\@namedef{resultnum@census/chat}{2{,}400}
\@namedef{resultnum@census/native}{1{,}500}
\@namedef{resultnum@census/unread}{52{,}000}
\@namedef{resultnum@census/models}{sixteen}
\@namedef{resultnum@census/models-cap}{Sixteen}
\@namedef{resultnum@census/complete-grid}{three}
\@namedef{resultnum@census/complete-grid-cap}{Three}
\@namedef{resultnum@pooled/mean}{$+7.7$}
\@namedef{resultnum@pooled/n-negative}{three}
\@namedef{resultnum@abstain/n-negative}{five}
\@namedef{resultnum@mass/breakpoint}{$0.480$ $[0.320, 0.520]$}
\@namedef{resultnum@dmask/abstain/n-excl-zero-neg}{four}
\@namedef{resultnum@recovery/mask}{$+19.5$}
\@namedef{resultnum@recovery-tool/mask}{$+32.5$}
\@namedef{resultnum@dlength/abstain/min}{$-20.0$}
\@namedef{resultnum@dlength/abstain/max}{$+0.0$}
\@namedef{resultnum@dlength/abstain/n-positive}{zero}
\@namedef{resultnum@dlength/abstain/n-excl-zero-neg}{four}
\@namedef{resultnum@recovery/length}{$-20.0$}
\@namedef{resultnum@recovery-tool/length}{$-12.5$}
\@namedef{resultnum@dtotal/abstain/min}{$-29.5$}
\@namedef{resultnum@dtotal/abstain/max}{$+1.5$}
\@namedef{resultnum@dtotal/abstain/n-positive}{one}
\@namedef{resultnum@dtotal/abstain/n-excl-zero-neg}{four}
\@namedef{resultnum@recovery/total}{$-0.5$}
\@namedef{resultnum@recovery-tool/total}{$+20.0$}
\@namedef{resultnum@recovery/abstain/free}{64.0}
\@namedef{resultnum@recovery/abstain/short}{44.0}
\@namedef{resultnum@recovery/abstain/constrained}{63.5}
\@namedef{resultnum@recovery/tool/free}{18.5}
\@namedef{resultnum@recovery/tool/short}{6.0}
\@namedef{resultnum@recovery/tool/constrained}{38.5}
\@namedef{resultnum@dmask/tool/n-positive-excl-zero}{six}
\makeatother
}{}

\maketitle
\draftbanner

\begin{abstract}
A tool-calling router has to pick the right tool when one applies and decline
when none does. Restricting the decoder to a grammar over the tool names is
the standard remedy for the first, and on small models it buys a large
accuracy gain.

Recent work separates the loss caused by asking for a format from the loss
caused by enforcing it at decode time. The second is small, which made
enforcement look nearly free. The same work declines to extend that to
function calling, where a constraint decides which answers exist rather than
how one is written. Declining to call anything is the answer it most easily
removes, and the one a router can least afford to lose.

A grammar decides which tokens may be emitted and where generation stops, and
a two-condition design charges both to the restriction. We therefore run three
conditions over one prompt: free generation, generation stopped at the first
line, and both applied together. We evaluate
open-weight models from 0.6B to 4B on the same items in English and Korean, comparing
the languages item by item.

The two-condition contrast is negative on abstention accuracy in
\resultnum{dtotal/abstain/n-excl-zero-neg} of
\resultnum{dmask/abstain/n-cells} cells with intervals excluding zero, and
positive in none, costing \resultnum{dtotal/abstain/min} points at worst. On
the smallest model in Korean the stop costs \resultnum{recovery/length}
points, the restriction returns
\resultnum{recovery/mask}, and together they leave
\resultnum{recovery/total}. What the restriction gives back is readable
output, not judgment. Of the \resultnum{repair/n-ft} abstentions it
repairs, \resultnum{repair/n-ft-unreadable} had no readable answer at all and
\resultnum{repair/n-ft-judgement} were correct decisions the scoring rule
rejected. On items that do need a tool the contrast is positive throughout,
and abstention is reported first because it is the registered measure. Both
preregistered claims about language fail: Korean does not lose more of the
abstentions it holds without the constraint, and the removed mass does not
explain what does.
\end{abstract}

\section{Introduction}
\label{sec:intro}

Declining is half of what a tool-calling router does. It has to pick the right
tool when one applies, and it has to call nothing when none does. The second
job keeps a deployed system safe, since a router that always calls something
will confidently invoke a weather API for a legal question. Constrained decoding is the standard remedy for the first.
Restricting the decoder to a grammar over the declared tool names removes
malformed output entirely, which on small models is worth a large accuracy
gain, and the intervention is close to free in engineering terms.

Recent work has also asked what it costs. \citet{formattax2026} decompose the
degradation that format constraints cause and find that most of it belongs to
the prompt rather than the decoder. Instructing a model to produce JSON costs
about $-3.9$ points on average; enforcing that instruction at decode time costs
about $-1.6$ points more. The reassuring reading, that enforcement is
nearly free once the instruction is paid for, is now the standard one.

That paper declines to extend the reading to function calling. Its
limitations section states that grammar-constrained decoding has other
applications (code synthesis, function calling, test generation) where
constraints encode correctness requirements rather than presentation
preferences, and that its results should not be extrapolated to those
settings.\footnote{\citet{formattax2026}, \S7. We take the exclusion at face
value and treat it as a scope boundary to be measured rather than as a
formality.} The distinction is exact and it is the opening this paper walks
through. A JSON schema restricts how an answer may be written. An enum over ten
tool names restricts which answers exist, and the answer it is most likely to
delete is the one that names no tool at all. If the decoder's contribution is
ever going to be large, it will be large here.

\paragraph{Separating the decoder from the prompt.}
A two-condition comparison cannot do it. Free generation against a single
constrained token differs in what the decoder may emit \emph{and} in how much
room the model has to answer, so the gap between them charges both to masking.
We therefore run three conditions over one byte-identical prompt: unconstrained
with 256 tokens to explain itself (\textbf{contract-free}), unconstrained but
stopped at the first line (\textbf{contract-short}), and enum-constrained on
the first line (\textbf{constrained}). Those are the names the rest of the
paper uses. Three conditions give three differences, and each answers a
different question:
\[
\begin{aligned}
  \Delta_{\text{mask}}(\ell)   &= \mathrm{acc}(\text{constr.}, \ell)
                                - \mathrm{acc}(\text{short}, \ell), \\
  \Delta_{\text{length}}(\ell) &= \mathrm{acc}(\text{short}, \ell)
                                - \mathrm{acc}(\text{free}, \ell), \\
  \Delta_{\text{total}}(\ell)  &= \mathrm{acc}(\text{constr.}, \ell)
                                - \mathrm{acc}(\text{free}, \ell).
\end{aligned}
\]
$\Delta_{\text{mask}}$ has one moving part, the grammar, but its baseline is
not neutral: the stop token is applied on both sides of it, so it is the cost
of adding an enum to a model whose answer length is already controlled rather
than the cost of constraining an otherwise free decoder. That second quantity
is $\Delta_{\text{total}}$, and the two differ by whatever the stop token
costs on its own, which Table~\ref{tab:delta} shows is not always small. We
report all three and never let one stand for another.

$\Delta_{\text{total}}$ is the one that lines up with prior work. A grammar
defines where generation stops as well as which tokens may be emitted, so the
constrained condition of \citet{formattax2026} carries a termination
constraint its unconstrained condition does not.\footnote{Their unconstrained
condition places no stop-sequence control on generation, which is what makes
\textbf{contract-free} rather than \textbf{contract-short} its counterpart
here.} Their decoder-side figure is already a sum of lexical masking and
termination measured against a baseline with neither, and separating those two
components is the methodological contribution of this paper.

\paragraph{Language, and what is borrowed.}
The masking argument predicts that the cost depends on the language, because
it depends on how much probability the model had placed outside the allowed
set before the mask. That quantity is the complement of the feasible mass of
\citet{dccd2026}, which we adopt as the registered mechanism and claim no part
of. Because our two languages are evaluated on the same items, the question is
asked within item: among the abstentions both languages get right under
\textbf{contract-short}, does the constraint break more of the Korean ones? It
does not, and the mass does not account for what does separate them, reported
as the negative result it is. We make no claim about absolute accuracy
anywhere; every reported quantity is a difference between conditions that saw
the same prompt.

\paragraph{Contributions.}
\begin{itemize}
  \item The size of the contrast prior work measures, in the
        function-calling abstention setting it set aside.
        $\Delta_{\text{total}}$ on abstention is negative in
        \resultnum{dtotal/abstain/n-excl-zero-neg} of
        \resultnum{dmask/abstain/n-cells} cells with intervals excluding zero,
        worst \resultnum{dtotal/abstain/min} points, and positive in none.
  \item That what the constraint recovers is repair rather than improvement.
        Of \resultnum{repair/n-ft} abstentions it turns from wrong to right,
        \resultnum{repair/n-ft-unreadable} had no readable answer to begin
        with and \resultnum{repair/n-ft-judgement} failed only on the
        scorer's ordering rule. Against free generation it also moves the
        decision boundary toward naming a tool.
  \item That the stop token and the enum are separable components with
        opposite signs, which a two-condition design sums into one number and
        which can cancel almost exactly: \resultnum{recovery/length},
        \resultnum{recovery/mask} and \resultnum{recovery/total} in one cell.
  \item A preregistered within-item test of whether the cost depends on the
        language, with the masked probability mass as the registered
        mechanism, and a breakpoint in that mass at
        \resultnum{mass/breakpoint} below which it predicts nothing about the
        per-item penalty and above which it predicts strongly.
  \item An audit of the MetaTool Task2 pairing the design rests on, covering
        four undocumented properties of the release, and a harness in which
        every table and figure is regenerated from the raw per-request records
        with an integrity check that fails the build on a number that cannot
        be reproduced from them.
\end{itemize}

The nearest antecedent to the second contribution is
\citet{zhou2026snowballing}, who enforces a reflection schema on a model
critiquing its own reasoning and finds near-perfect surface conformance with
the deeper errors undetected. Surface form is bought and judgment is not,
which is what this paper reports as well. He attributes it to cognitive load,
offered as an explanation rather than measured; counting what each repaired
output looked like before the constraint turns that hypothesis into
arithmetic. Appendix~\ref{sec:appendix-antecedents} sets the two designs side
by side.

%
%
\section{Related Work}
\label{sec:related}

\paragraph{A constraint tax is now measured in five places, and abstention is
not one of them.}
\citet{speakfreely2024} showed that requiring a format degrades task
performance, and four papers have since taken that apart. \citet{formattax2026}
decomposed the degradation by where it enters, finding most of it in what the
format instruction does to the prompt (mean $-3.9$ points) and little in the
decoder enforcing it (mean $-1.6$), which is the conclusion that made
enforcement look nearly free. \citet{ray2026constrainttax} found schema
validity rising to complete at sub-3B scale while answer accuracy falls.
\citet{zhou2026snowballing} found that a reflection schema buys surface
conformance without improving self-correction. \citet{li2026suppression} found
that a JSON Schema constraint stops open-weight models calling tools at all.

The five cover reasoning accuracy, schema validity, self-correction and tool
invocation. None measures whether a model still declines when declining is the
right answer, which is the behavior an enum over tool names is most likely to
remove and the one a deployed router is least able to lose.

\paragraph{The mechanism is formalized.}
\citet{dccd2026} cast masking-and-renormalization as a KL projection onto the
feasible set and show that the distortion is governed by the feasible mass,
how much probability the model had placed inside the allowed set before the
mask. The quantity this paper measures per item, \texttt{masked\_mass}, is one
minus that: their construct applied to a tool-calling decoder, and we claim no
part of it. \citet{gidle2025} gives the nearest empirical handle on the same
mechanism in Korean, banning another script's tokens in open-ended generation
and finding naive masking more variable than a distortion-minimizing
renormalization, without measuring the mass itself.

There is a line of work on correcting that distortion rather than measuring
it. \citet{park2024gad} named the problem, that a constrained decoder returns
grammatical output whose likelihoods are not proportional to the ones the
model assigned; \citet{loula2025smc} extended the correction to sequential
Monte Carlo, and \citet{dang2026lcdbias} attacks the same distortion from the
other end, building globally constrained proposals rather than masking the
next token myopically.

We measure where the bias they correct becomes large enough to see in an
outcome. The relationship between the masked mass and the per-item penalty is
a hinge rather than a line, flat below a breakpoint at
\resultnum{mass/breakpoint} and steep above it
(\S\ref{sec:appendix-exploratory}), so a correction to myopic masking has
something to bite on above that point and little below it. We do not test that
prediction here.

\paragraph{Function calling was declared out of scope, and abstention is where
that boundary bites.}
\citet{formattax2026} exclude function calling in their \S7, warning against
extrapolating to settings where a constraint encodes a correctness requirement
rather than a presentation preference. Their own boundary marks the case where
the decoder's contribution should be largest, and it is the case a deployed
router runs in. One paper reaches it from the other side:
\citet{lee2026patool}, adapting tool schemas to small models, report that an
enum constraint raising tool-selection accuracy lowers accuracy on the items
whose correct answer is to abstain. That measurement is English-only.

\paragraph{Neither line crosses language.}
\citet{luo2026lostexecution} give the most recent diagnosis of multilingual
tool calling, where the dominant failure is parameter value language mismatch:
the model picks the right tool, then writes argument values in the user's
language and breaks a contract that expects English. What matters here is the
layer. All three mitigations they evaluate leave the decoder alone, by design,
since the point is an intervention that drops into a deployed system. Their
evaluation is built on the multiple-function subset of BFCL, which offers no
item whose correct answer is to call nothing, and their languages are Chinese,
Hindi and Igbo.

Korean tool calling has been approached from the opposite direction.
\citet{garg2026luckystar} adapt a 111B model with multilingual supervised
fine-tuning and reinforcement learning with verifiable rewards. That is
training-time adaptation of a large model where this paper is inference-time
constraint of small ones, and it too has no abstention axis. Between the two,
the sub-4B range, where constrained decoding is most attractive precisely
because the base model is least reliable, is unexamined outside English, and
what a decoder-level constraint does to abstention is unexamined in any
language.

\paragraph{Two of the five are close enough to need separating.}
\citet{li2026suppression} is the sharpest result in the cluster and the cause
is mechanical: the schema compiles to an automaton in which the bracket that
opens a Qwen-style \texttt{<tool\_call>} tag is forbidden in every state, so
the behavior it would introduce cannot occur. Their constraint and the
behavior it suppresses lie on separate axes, while ours is an enum over the
answer set itself with the abstention option inside it by construction. Their
outcome is uniform across models, as a mechanical consequence should be; ours
is not, because $\Delta_{\text{mask}}$ changes sign from cell to cell.
\citet{ray2026constrainttax} is the closer of the two in scope and is where
the phrase constraint tax comes from, but he measures no abstention, has no
second language, and does not decompose the constraint, so a tax is a tax
everywhere he looks. Appendix~\ref{sec:appendix-antecedents} works both
comparisons through.

\paragraph{This paper measures the intersection.}
Tool abstention, under an enum constraint, in two languages, on models small
enough that the constraint is standard practice, with the decoder's
contribution separated from the prompt's by a third condition that changes
output length without changing the grammar. We use MetaTool
\citep{metatool2024}, whose Task2 provides matched tool-needed and abstention
items over a shared query set, with the scoring rules of \citet{lee2026patool}
rather than MetaTool's own. Hypotheses, falsification conditions and analysis
were fixed before collection \citep{prereg2021}, and the build fails if the
registered hypothesis section is edited after results exist.

\section{Experimental Setup}
\label{sec:setup}

\subsection{Items}

We build on MetaTool \citep{metatool2024} Task2, which contains two files of
995 rows each. Subtask1 supplies items where a suitable tool exists among the
candidates; Subtask3 supplies items where it does not. The two files are
\emph{positionally paired}: row $i$ of each carries the same user query and the
same value in the \texttt{tool} field. We verified this over all 995 rows
(query 995/995, \texttt{tool} 995/995), and every row of both files offers
exactly ten candidate tools.

Sampling therefore selects \emph{pair positions}, never rows from one file
independently. Two hundred positions are drawn once, without replacement, under
a fixed seed, and both halves of each selected pair enter the evaluation set,
giving 200 abstention items and 200 tool-needed items whose queries, lengths
and vocabulary are matched by construction. No item is filtered for difficulty.

\paragraph{The gold label for abstention items.}
For a Subtask3 item the correct answer is the literal string \texttt{None}. The
\texttt{tool} field of that row names the tool that was \emph{removed} from the
candidate list; it is not the answer. We confirmed the removal holds for all
995 rows: the \texttt{tool} value appears among the ten candidates in 995/995
Subtask1 rows and in 0/995 Subtask3 rows. Treating that field as the gold label
would invert the measurement, and the released harness raises an error
rather than allowing it.

\paragraph{Item identity.}
The released \texttt{index} field is not usable for alignment: Subtask3 stores
a single repeated value across all 995 rows, and the two files agree on it for
one row. Items are therefore identified by list position.

\paragraph{Properties of the released data.}
Four properties of the release bear on interpretation and we did not find them
documented: the ten candidates offered to a paired item are disjoint between
the two files, the few-shot block differs systematically by item type, the user
query appears twice in each prompt, and the candidate list is ordered rather
than shuffled, with the gold tool first in all 995 Subtask1 rows. None of them
disturbs a within-item difference between conditions that see the same prompt,
and none can manufacture abstention accuracy, since the abstention answer is
not in the tool list at any position. \S\ref{sec:appendix-release} states each
one and what it does and does not threaten, and Limitations carries them
forward.

\subsection{Conditions}

Three conditions, one prompt. The MetaTool completion prompt is used verbatim
in all three; it already instructs the model to return \texttt{'None.'} when no
tool applies, so abstention is an explicitly offered option throughout. What
differs is the decoding.

\textbf{contract-free.} Unconstrained, with a deliberately generous budget of
256 tokens, because truncating a justification inflates the apparent cost of a
constraint.

\textbf{contract-short.} Unconstrained, stopped at the first newline. The model
answers on one line and does not explain itself.

\textbf{constrained.} The same one-line budget, with the decoder restricted to
a choice over the item's ten candidate names \emph{plus} \texttt{None}.
Including the abstention option is essential: a grammar that omits it makes
abstention physically impossible and the comparison vacuous. What we measure is
whether a model abstains less \emph{when it still can}.

The middle condition is what makes the design a measurement. Between
\textbf{contract-short} and \textbf{constrained} the request differs in
exactly one field, the grammar, and that difference is
$\Delta_{\text{mask}}$; between \textbf{contract-free} and
\textbf{contract-short} lies the cost of taking away the room to answer,
$\Delta_{\text{length}}$, which a two-condition design would charge to the
constraint. Their sum is what such a design would report.

$\Delta_{\text{mask}}$ attributes an effect to the grammar and nothing else,
against a baseline that is already format-controlled, so it is the marginal
cost of an enum on top of a stop token. $\Delta_{\text{total}}$ compares the
constrained decoder against a decoder left alone, which is the comparison
prior work makes and the one a practitioner switching constrained decoding on
is facing. The two coincide only where $\Delta_{\text{length}}$ is zero, and
\S\ref{sec:results} shows it is not. The harness builds all three requests
before a sweep and refuses to run if their prompts are not byte-identical.

All three run at temperature 0 across five server seeds, on Qwen3 0.6B, 1.7B
and 4B, in English and Korean: $3 \times 2 \times 3 \times 5 \times 400 =
\resultnum{census/confirmatory}$ requests. That is the registered analysis, and
it is the number every confirmatory claim in this paper rests on.

The full instrument list, the sampling parameters, the retry policy and the
records kept alongside the primary measurement are in
\S\ref{sec:appendix-collected}.

\subsection{Korean localization}

The user query is translated; tool names, descriptions, instructions and
few-shot examples remain in English. This is the deployed condition of
interest, a language-invariant execution interface driven by a non-English
user. Both occurrences of the query are replaced, and the harness asserts that
every other byte is identical to the English counterpart. Machine translation
is screened by back-translation, items over the drift threshold are excluded
with a recorded reason, and evaluation is blocked until a human has reviewed
the translations.

\subsection{Scoring}

We adopt the single-match and ordering rules of \citet{lee2026patool}. For an
abstention item a correct answer contains the abstention token and either no
candidate mention or one that follows it; for a tool-needed item it mentions
exactly the gold tool, ahead of any abstention token. We depart from them in
one place: where an output names two or more candidates, their procedure asks
an LLM judge which was selected and we score it wrong and count it, keeping
the pipeline deterministic. That makes us the stricter of the two, and the
multi-candidate counts are reported alongside the accuracies rather than
folded into them. \S\ref{sec:appendix-scoringdetail} gives the departure in
full and the word-boundary matching the candidate pool requires.

\subsection{Mechanism variable}

In the unconstrained conditions we record the top-$k$ first-token distribution
and sum the probability mass on tokens the enum grammar would have forbidden,
which is the complement of the feasible mass of \citet{dccd2026}. Allowed
tokens are derived from the tokenizer for every enum member, admitting both
space-prefixed and bare encodings; admitting extra spellings can only shrink
the quantity, and because $k$ is finite the value is a lower bound whose
tightness is reported alongside it. The analysis reads the measurement from
\textbf{contract-short}, the control condition of the primary contrast.

\section{Results}
\label{sec:results}

Table~\ref{tab:main} reports abstention accuracy and tool-selection accuracy
for every model, language and condition. The two are always shown side by side:
abstention accuracy alone is maximized by a model that answers \texttt{None}
unconditionally, and the tool-selection column is what exposes that
degenerate strategy.

\begin{table*}[t]
\centering
\small
\setlength{\tabcolsep}{3pt}
\begin{tabular}{llrrrrrr}
\toprule
Model & Lang & Free abstain & Free tool sel. & Short abstain & Short tool sel. & Constrained abstain & Constrained tool sel. \\
\midrule
Qwen3-0.6B & English & 65.0 \tiny{[58.5, 71.5]} & 26.0 \tiny{[20.0, 32.5]} & 59.0 \tiny{[52.0, 66.0]} & 16.5 \tiny{[11.5, 22.0]} & 35.5 \tiny{[28.5, 42.5]} & 43.0 \tiny{[36.0, 50.0]} \\
 & Korean & 64.0 \tiny{[57.0, 70.5]} & 18.5 \tiny{[13.5, 24.0]} & 44.0 \tiny{[37.0, 51.0]} & 6.0 \tiny{[3.0, 9.5]} & 63.5 \tiny{[57.0, 70.0]} & 38.5 \tiny{[32.0, 45.5]} \\
\midrule
Qwen3-1.7B & English & 77.0 \tiny{[71.0, 82.6]} & 33.8 \tiny{[27.5, 40.5]} & 74.5 \tiny{[68.0, 80.5]} & 34.0 \tiny{[27.5, 40.5]} & 50.5 \tiny{[43.5, 57.5]} & 45.0 \tiny{[38.0, 52.0]} \\
 & Korean & 73.0 \tiny{[66.9, 78.9]} & 30.9 \tiny{[24.7, 37.2]} & 61.5 \tiny{[54.5, 68.0]} & 30.5 \tiny{[24.5, 37.0]} & 48.5 \tiny{[41.5, 55.0]} & 41.0 \tiny{[34.0, 48.0]} \\
\midrule
Qwen3-4B & English & 57.0 \tiny{[50.1, 63.8]} & 12.9 \tiny{[8.6, 17.5]} & 57.0 \tiny{[50.1, 63.8]} & 66.6 \tiny{[60.0, 73.1]} & 48.2 \tiny{[41.2, 55.0]} & 72.5 \tiny{[66.0, 78.5]} \\
 & Korean & 58.2 \tiny{[51.3, 64.9]} & 6.0 \tiny{[3.1, 9.6]} & 57.7 \tiny{[50.8, 64.4]} & 55.2 \tiny{[48.1, 62.1]} & 59.7 \tiny{[53.0, 66.4]} & 68.7 \tiny{[62.2, 75.0]} \\
\bottomrule
\end{tabular}
\caption{Abstention accuracy and tool-selection accuracy (\%) with 95\% bootstrap confidence intervals. Intervals resample items and seeds jointly. Absolute values are reported for context only; the design licenses claims about the differences in Table~\ref{tab:delta}, not about these levels.}
\label{tab:main}
\end{table*}

Table~\ref{tab:paired} reports the quantity the paper is about: the registered
within-item statistic of \S\ref{sec:analysis}, which asks whether the
constraint breaks more Korean abstentions than English ones on the items both
languages held without it. Because both languages enter that conditional set
at the same accuracy by construction, it cannot be moved by a difference in
baselines, the defect that retired the absolute difference before any of this
data existed. That absolute difference is in the same table and the retention
ratio is in \S\ref{sec:appendix-metric}; where the three disagree, the
disagreement is reported rather than resolved by choosing one.

\begin{table*}[t]
\centering
\small
\begin{tabular}{lrr rrrr}
\toprule
Model & $|S|$ & KO / EN only & $D_{\text{paired}}$ & Short English & Short Korean & $\Delta$(KO) $-$ $\Delta$(EN) \\
\midrule
Qwen3-0.6B & 375 & 15 / 100 & \textbf{-22.7 \tiny{[-34.2, -11.4]}} & 59.0 \tiny{[52.0, 66.0]} & 44.0 \tiny{[37.0, 51.0]} & \textbf{+43.0 \tiny{[+33.5, +52.5]}} \\
Qwen3-1.7B & 550 & 65 / 50 & +2.7 \tiny{[-5.7, +11.3]} & 74.5 \tiny{[68.0, 80.5]} & 61.5 \tiny{[54.5, 68.0]} & \textbf{+11.0 \tiny{[+2.5, +19.5]}} \\
Qwen3-4B & 480 & 4 / 59 & \textbf{-11.5 \tiny{[-18.5, -4.9]}} & 57.0 \tiny{[50.1, 63.8]} & 57.7 \tiny{[50.8, 64.4]} & \textbf{+10.8 \tiny{[+4.8, +16.8]}} \\
\bottomrule
\end{tabular}
\vspace{2pt}
\begin{minipage}{\linewidth}\footnotesize The \emph{Short} columns are \textbf{contract-short}, not \textbf{contract-free}: both are unconstrained, and the difference between them is the length term this paper reports separately. $|S|$ is the size of that conditional set and the next column gives the discordant counts $b$ and $c$; items both languages lost, or both held, carry no information about which language suffers more and cancel. \textbf{H2 predicts a positive $D_{\text{paired}}$}: it counts failures, so its sign is the reverse of an accuracy difference, where H2 predicts negative. The two middle columns give the level each language starts from, unconstrained, because every difference here is bounded by it. The retention ratio is reported in \S\ref{sec:appendix-metric} instead: it divides by that level, and where the level is low the ratio leaves the unit interval. The final column is the metric retired before the confirmatory data existed, kept because a disagreement between the measures is itself a result.\end{minipage}
\caption{The registered primary statistic and the two secondary measures of the same effect. $D_{\text{paired}}$ is the McNemar contrast on the items both languages abstained on correctly without the constraint: the share only Korean lost minus the share only English lost. \textbf{Bold} marks an interval excluding zero.}
\label{tab:paired}
\end{table*}

Table~\ref{tab:delta} reports all three differences the design makes
available, for both item types and both languages. They should be read
together, because two of them can be large and still cancel.

The clearest case is the smallest model in Korean. Stopping the answer at the
first line costs \resultnum{recovery/length} points of abstention accuracy;
adding the enum on top returns \resultnum{recovery/mask}; the two together
leave \resultnum{recovery/total}, which is nothing. Read alone,
$\Delta_{\text{mask}}$ there says the grammar manufactures abstentions. Read
beside $\Delta_{\text{length}}$ it says something narrower and more
defensible: the grammar gives back what a different piece of format control
had just taken away, and does not carry the model past where it started.

\paragraph{The recovered abstentions are not a degenerate strategy.}
A model that has moved probability mass onto the abstention token will abstain
more, and on abstention items that is indistinguishable from abstaining
better. What separates the two is the other half of the benchmark, and here
the two move together: the constraint takes tool selection up by
\resultnum{recovery-tool/mask} points in the same cell in which it takes
abstention up by \resultnum{recovery/mask}. A model buying abstention by
defaulting to \texttt{None} would have to give up the tool-needed items, and
this one gains them. That holds throughout: on tool-needed items the
constraint raises accuracy in
\resultnum{dmask/tool/n-positive-excl-zero} of
\resultnum{dmask/abstain/n-cells} cells with intervals excluding zero and
lowers it in none, and Table~\ref{tab:retention} in
\S\ref{sec:appendix-metric} puts both accuracies side by side cell by cell.

Both columns rising together \emph{against the stop-token baseline} is what a
repair account predicts and not what an improvement account would. Output with
no readable answer is unreadable whether the item needed a tool or needed
none, so a grammar that makes output readable acts on both halves at once,
while a grammar that improved judgment would have to improve two different
decisions. The claim is specific to $\Delta_{\text{mask}}$: against the free
baseline the two columns do not move together at all
(\S\ref{sec:results-pooled}).

\paragraph{Nor is it the mirror strategy.}
The gold tool is the first candidate in every row of this benchmark
(\S\ref{sec:setup}), so a decoder pulling the model toward the head of the
list would raise tool-selection accuracy with no improvement in judgment. The
pull is small on the half where naming the first candidate is always wrong:
the share of abstention outputs naming it moves by
\resultnum{position/abstain-shift-mean} points on average between the
unconstrained and constrained conditions. A companion note permutes the
ordering and finds the shipped one is a headwind rather than a tailwind
\citep{positionbias2026}, and in any case the ordering is byte-identical
across the three conditions, so it cancels out of every difference reported
here. Appendix~\ref{sec:appendix-ordering} gives the full argument.

The pattern holds across the confirmatory cells. Against the unconstrained,
unstopped baseline, $\Delta_{\text{total}}$ on abstention runs from
\resultnum{dtotal/abstain/min} to \resultnum{dtotal/abstain/max} points. It is
negative in \resultnum{dtotal/abstain/n-excl-zero-neg} of
\resultnum{dmask/abstain/n-cells} cells with intervals excluding zero, and
positive in none. Against the
stop-token baseline, $\Delta_{\text{mask}}$ runs from
\resultnum{dmask/abstain/min} to \resultnum{dmask/abstain/max} and is positive
in \resultnum{dmask/abstain/n-positive}. The difference between those two
readings is $\Delta_{\text{length}}$, which is negative with an interval
excluding zero in \resultnum{dlength/abstain/n-excl-zero-neg} cells and
positive in \resultnum{dlength/abstain/n-positive}.

\begin{table*}[t]
\centering
\small
\begin{tabular}{lllrrr}
\toprule
Model & Metric & Contrast & $\Delta$ English & $\Delta$ Korean & $\Delta$(KO) $-$ $\Delta$(EN) \\
\midrule
Qwen3-0.6B & Abstention & Masking (constrained $-$ short) & -23.5 \tiny{[-32.0, -15.5]} & +19.5 \tiny{[+10.5, +28.5]} & \textbf{+43.0 \tiny{[+33.5, +52.5]}} \\
Qwen3-0.6B & Abstention & Length (short $-$ free) & -6.0 \tiny{[-9.5, -3.0]} & -20.0 \tiny{[-25.5, -14.5]} & \textbf{-14.0 \tiny{[-20.5, -7.5]}} \\
Qwen3-0.6B & Abstention & Both (constrained $-$ free) & -29.5 \tiny{[-37.5, -21.5]} & -0.5 \tiny{[-9.5, +8.0]} & \textbf{+29.0 \tiny{[+19.0, +39.0]}} \\
\midrule
Qwen3-0.6B & Tool selection & Masking (constrained $-$ short) & +26.5 \tiny{[+19.0, +34.0]} & +32.5 \tiny{[+25.5, +39.5]} & +6.0 \tiny{[-1.5, +13.5]} \\
Qwen3-0.6B & Tool selection & Length (short $-$ free) & -9.5 \tiny{[-16.5, -3.0]} & -12.5 \tiny{[-18.0, -7.5]} & -3.0 \tiny{[-11.5, +5.5]} \\
Qwen3-0.6B & Tool selection & Both (constrained $-$ free) & +17.0 \tiny{[+9.5, +24.5]} & +20.0 \tiny{[+12.0, +28.0]} & +3.0 \tiny{[-6.0, +12.0]} \\
\midrule
Qwen3-1.7B & Abstention & Masking (constrained $-$ short) & -24.0 \tiny{[-32.0, -16.0]} & -13.0 \tiny{[-21.5, -4.5]} & \textbf{+11.0 \tiny{[+2.5, +19.5]}} \\
Qwen3-1.7B & Abstention & Length (short $-$ free) & -2.5 \tiny{[-5.0, -0.5]} & -11.5 \tiny{[-16.0, -7.3]} & \textbf{-9.0 \tiny{[-13.5, -4.5]}} \\
Qwen3-1.7B & Abstention & Both (constrained $-$ free) & -26.5 \tiny{[-34.4, -18.8]} & -24.5 \tiny{[-32.5, -16.5]} & +2.0 \tiny{[-6.8, +11.0]} \\
\midrule
Qwen3-1.7B & Tool selection & Masking (constrained $-$ short) & +11.0 \tiny{[+5.0, +17.0]} & +10.5 \tiny{[+3.0, +18.0]} & -0.5 \tiny{[-8.0, +7.0]} \\
Qwen3-1.7B & Tool selection & Length (short $-$ free) & +0.2 \tiny{[-5.0, +5.5]} & -0.4 \tiny{[-6.9, +6.0]} & -0.6 \tiny{[-8.6, +7.3]} \\
Qwen3-1.7B & Tool selection & Both (constrained $-$ free) & +11.2 \tiny{[+4.0, +18.0]} & +10.1 \tiny{[+2.0, +18.1]} & -1.1 \tiny{[-9.6, +7.3]} \\
\midrule
Qwen3-4B & Abstention & Masking (constrained $-$ short) & -8.8 \tiny{[-13.7, -4.2]} & +2.0 \tiny{[-2.4, +6.5]} & \textbf{+10.8 \tiny{[+4.8, +16.8]}} \\
Qwen3-4B & Abstention & Length (short $-$ free) & +0.0 \tiny{[+0.0, +0.0]} & -0.5 \tiny{[-1.5, +0.0]} & -0.5 \tiny{[-1.5, +0.0]} \\
Qwen3-4B & Abstention & Both (constrained $-$ free) & -8.8 \tiny{[-13.7, -4.2]} & +1.5 \tiny{[-3.0, +6.1]} & \textbf{+10.3 \tiny{[+4.2, +16.5]}} \\
\midrule
Qwen3-4B & Tool selection & Masking (constrained $-$ short) & +5.9 \tiny{[+1.5, +10.4]} & +13.5 \tiny{[+8.3, +18.9]} & \textbf{+7.6 \tiny{[+2.0, +13.4]}} \\
Qwen3-4B & Tool selection & Length (short $-$ free) & +53.7 \tiny{[+46.4, +61.0]} & +49.2 \tiny{[+42.0, +56.4]} & -4.5 \tiny{[-11.6, +2.5]} \\
Qwen3-4B & Tool selection & Both (constrained $-$ free) & +59.6 \tiny{[+52.7, +66.5]} & +62.7 \tiny{[+56.0, +69.5]} & +3.1 \tiny{[-3.0, +9.2]} \\
\bottomrule
\end{tabular}
\vspace{2pt}
\begin{minipage}{\linewidth}\footnotesize The masking row is the effect this paper is about: the grammar applied with the answer length already fixed. The length row is what shortening the answer costs on its own, and a two-condition design charges it to the constraint. The total row is their sum, which is what such a design would report. Intervals are percentile bootstrap ($B = 10000$) resampling items and seeds jointly, with the same draw in every cell of a contrast.\end{minipage}
\caption{Each difference the three conditions make available, in accuracy points. Negative values mean the change hurt. \textbf{Bold} marks an interval that excludes zero.}
\label{tab:delta}
\end{table*}

\subsection{Both item types together}
\label{sec:results-pooled}

Table~\ref{tab:pooled}, in \S\ref{sec:appendix-pooled}, pools the two
equinumerous item types. \textbf{It does not have the same sign as the
abstention column}: the mean over cells is \resultnum{pooled/mean} points,
against \resultnum{dtotal/abstain/min} at worst on abstention alone. Every
headline here is the abstention column, because that is what the
preregistration names as the primary measure, and the pooled column is stated
by us rather than left to be discovered. The pooled number being kinder to the
intervention makes moving to it worse rather than better.

Against the free baseline the two columns move apart: tool selection rises in
every cell while abstention falls in \resultnum{abstain/n-negative} of
\resultnum{dmask/abstain/n-cells}. That is a decision threshold moving toward
naming a tool, and it is a different thing from the repair above. Measured
from the stop token the grammar gives back what that stop token destroyed and
both columns rise; measured from free generation it also shifts where the
model draws the line between acting and declining, which pays on the tool half
and costs on the abstention half. The shift is not uniform, and
Appendix~\ref{sec:appendix-pooled} reports how it differs by model. One
intervention that a two-condition design reports as a single number is doing
two separable things with opposite signs, and reporting them apart is the
point of the third condition.

\section{Analysis}
\label{sec:analysis}

\subsection{Confirmatory test}

The preregistered test is a single comparison, made within items. Let $S$ be
the abstention items that both languages answered correctly under
\textbf{contract-short}, and let a \emph{flip} be such an item whose abstention
the constraint then breaks. With $b$ the flips that occur only in Korean and
$c$ those that occur only in English,
\[
  D_{\text{paired}} = \frac{b - c}{|S|},
\]
reported in Table~\ref{tab:paired}. H2, the registered language
hypothesis, holds that the constraint breaks more of the Korean abstentions in
$S$ than of the English ones, and so predicts a positive value: $D$ counts
failures, not accuracy. The registered falsification condition is that the 95\%
confidence interval of this quantity contains zero, and an empty discordant set
($b + c = 0$) falsifies as well, there being nothing in the data that
distinguishes the languages.

Conditioning on $S$ is what makes the comparison a comparison. Abstention
accuracy has a ceiling, so a language that starts higher has further to fall,
and an absolute difference of differences carries that gap inside it. Both
languages enter $S$ at the same accuracy by construction, which is available
here only because the design puts them on the same items
(\S\ref{sec:appendix-metric}).

Intervals are percentile bootstrap over 10{,}000 replicates resampling
evaluation items and decoding seeds jointly. Resampling items alone
understates uncertainty for the smaller models, whose accuracy moves between
seeds even at temperature 0 because the server seed changes the constrained
decoder's tie-breaking. Within a replicate the same resampled items and seeds
are used in every cell of a contrast, so differences are paired.

\subsection{Mechanism}

The second registered analysis, H3, asks whether the mass accounts for the
language: whether the mass predicts an item's flip, and whether adding the
language to that model explains anything the mass has not.
The unit is one observation per item, language and seed, over the same
conditional set. We fit
\[
  \Pr(\text{flip}) = \sigma(\beta_0 + \beta_1\,\texttt{masked\_mass})
\]
and the same model with a language term added, reported in
Table~\ref{tab:mechanism} in \S\ref{sec:appendix-massfigs}. If the language
coefficient's interval covers zero once the mass is in the model, language
acts on abstention through the mass rather than beside it; if it excludes
zero, something is left that the mass does not explain. The analysis is
registered as primary in its own right and is reported whatever
$D_{\text{paired}}$ turns out to be.

\S\ref{sec:appendix-massfigs} carries the regression, the coverage table and
three figures: the per-item penalty against \texttt{masked\_mass} by language
(Figure~\ref{fig:mechanism}), the mass distributions themselves
(Figure~\ref{fig:mass}), and the penalty against parameter count
(Figure~\ref{fig:size}).

Because \texttt{masked\_mass} is summed over a truncated head of the
distribution it is a lower bound; Table~\ref{tab:mass}, in
\S\ref{sec:appendix-massfigs}, reports how much of the distribution that head
covered, which is what makes the bound interpretable.

\subsection{Model size}

Whether the penalty grows as models shrink is registered as a secondary
observation. Figure~\ref{fig:size}, in \S\ref{sec:appendix-massfigs}, plots
$\Delta_{\text{abstain}}$ against parameter count for each language.

\subsection{Analyses not registered in advance}

Any analysis added after seeing the data is reported in
\S\ref{sec:appendix-exploratory} and labeled exploratory. It carries no
confirmatory weight, and the registered test above is unaffected by it.

\section{Conclusion}
\label{sec:conclusion}

Constrained decoding costs abstention accuracy in the setting prior work set
aside, and the result that made enforcement look nearly free does not hold
there. A single before-and-after number also hides that the intervention is
two things at once. A grammar fixes where generation stops as well as which
tokens may be emitted, and on abstention items those two effects have opposite
signs and can cancel almost exactly. Separating them is what the third
condition is for.

What the grammar buys is readable output. The abstentions it turns from wrong
to right were almost all outputs with no readable answer in them, and none
were correct decisions the scoring rule had rejected, so it is not releasing
judgments the model had already made. Measured against free generation it also
moves the decision boundary toward calling a tool. Neither of those is better
judgment.

Three things are open, and the Limitations section takes up each of them. Our
account and a headroom explanation fit the cell-level regression equally well.
The decomposition rests on six cells, because \textbf{contract-free} was
collected in full for three models only. And both preregistered claims about
language fail, which leaves the support we had for expecting a language effect
thinner than it looked.

For practice the consequence is narrow and firm. Report the stop rule and the
grammar separately, because one number can be a large loss and a large
recovery reported as nothing. And do not expect a grammar to improve a
decision. It makes unreadable output readable, which is worth having, and it
leaves the judgment as it was.

\section*{Limitations}
\label{sec:limitations}
\addcontentsline{toc}{section}{Limitations}

\paragraph{The gold tool is always the first candidate offered.}
The released candidate lists are ordered, not shuffled, and the gold tool is
first in every Subtask1 row (\S\ref{sec:setup}). Naming the first candidate
without reading the query therefore answers the tool-needed half perfectly, so
a tool-selection score on this benchmark is not by itself evidence of tool
selection. Our own data cannot separate the two, because the subset that would
do it, items whose gold is not first, is empty.

A companion note runs the permutation the benchmark does not ship
\citep{positionbias2026}. Moving the gold through all ten positions, in English
and on four models, it finds that none of them answers by naming the first
candidate, and that accuracy is in fact lower when the gold is first than when
it is elsewhere. So the ordering costs these models accuracy rather than buying
it, and the tool-selection levels reported here are a lower bound. That evidence
bounds the concern rather than removing it: it covers one language and four
models, on permutations of the benchmark rather than the benchmark itself.

Either way the concern reaches levels and not contrasts. The ordering is
identical across the three conditions, so it cancels out of every difference
this paper reports.

\paragraph{Only three models have all three conditions.}
\textbf{contract-free} was collected in full for the three registered models
and for no others (\S\ref{sec:setup}). $\Delta_{\text{length}}$ and
$\Delta_{\text{total}}$ therefore exist for six cells, and the decomposition
that this paper is built around cannot be checked on the other thirteen models.
What those models support is the single contrast that was collected for all of
them, and the exploratory analyses say so.

\paragraph{At cell level the account is not distinguishable from headroom.}
Splitting the baseline's error into its unreadable and readable halves gives
coefficients that cannot be told apart, so the two-term model reduces to the
baseline model. The constraint's benefit then tracks how much the baseline got
wrong rather than what kind of wrong it was (\S\ref{sec:appendix-cells}). The
item-level transition counts are not subject to that objection, because they
condition on the repairs that happened and ask what those outputs were
beforehand, but the cell-level regression does not support the reading and we
do not claim that it does.

\paragraph{Distractor sets are not matched, and priming is not either.}
The two halves of a pair share a query and a gold \texttt{tool} field but offer
disjoint candidate sets, so the pairing controls query difficulty and not
distractor difficulty. Every abstention prompt also carries two few-shot
examples answered \texttt{None} where every tool-needed prompt carries none.
Both are differenced out of a within-item contrast, since both conditions see
the identical prompt, but they mean the two columns of Table~\ref{tab:main} are
not a like-for-like comparison and that absolute abstention accuracy is
inflated relative to a neutrally primed prompt.

\paragraph{Abstention is scored from a mention rule, not from an intent.}
The rules resolve position, not meaning, so a candidate name that is also an
ordinary English word can be matched inside a justification that was not naming
a tool; \texttt{Now} is the clearest case. We restrict matching to the item's
own ten candidates and require word boundaries, which removes every
name-inside-name collision in this pool, but the residual ambiguity is real and
is quantified in the released collision report. Where an output names two or
more candidates we score it wrong, whereas the rules we adopt it from refer
that case to an LLM judge (\S\ref{sec:setup}).

\paragraph{The mechanism variable is a lower bound.}
\texttt{masked\_mass} is summed over a truncated top-$k$ head, so it
understates the true masked mass by the unobserved tail, and it is computed
from first tokens only, so it cannot capture constraint pressure appearing
later in a name.

\paragraph{The Korean queries are translated and reviewed, not native.}
Only the user query is translated; tool names, descriptions, instructions and
few-shot examples are the English original byte for byte, which is the deployed
condition of interest. Queries were drafted and then read in full by a native
speaker who could correct, accept or exclude each one, and both the discarded
first draft and the reviewed set are released. Two things limit the distortion:
the reported quantity differences two conditions that see the identical prompt,
so whatever translation does to a query it does to both; and the language
comparison is within item, on items both languages answered correctly
unconstrained. What neither controls is a systematic shift in how Korean
queries are phrased, and for that a natively authored control set is released
and analyzed separately (\S\ref{sec:appendix-schema}); it is underpowered and
excludes nothing.

\paragraph{Arguments are out of reach, and so is a second stack.}
MetaTool Task2 items have no argument fields, so the parameter value language
mismatch of \citet{luo2026lostexecution} cannot be measured here at all; the
preregistration lists it as a secondary observation and no substitute has been
improvised.
All results also come from one inference stack with one grammar backend, and
constraint behavior differs between grammar engines, so these numbers should
not be carried to a different stack without re-measurement.

\section*{Ethics Statement}
\label{sec:ethics}
\addcontentsline{toc}{section}{Ethics Statement}

The evaluation data is the public MetaTool benchmark (MIT). No human subjects
were involved and no personal data was processed. The work evaluates whether a
common deployment technique degrades a model's ability to decline to act, which
we regard as a safety-relevant failure mode; the release includes the full
per-request records so the claim can be checked rather than taken on trust.

\bibliography{refs}

\appendix
\section{Reproduction}
\label{sec:appendix-repro}

The release contains the harness, the raw per-request records and the
generators for every table and figure. \texttt{make verify} regenerates all
generated artifacts into a scratch directory and fails if any committed
artifact differs, so no number in this manuscript can drift from the records it
came from.

\begin{table}[h]
\centering\footnotesize
\begin{tabular}{@{}l p{0.58\columnwidth}@{}}
\toprule
Target & Effect \\
\midrule
\texttt{make check}   & SSH reachability and GPU inventory \\
\texttt{make smoke}   & 10 pairs, both languages, both conditions \\
\texttt{make dataset} & manifest, translation draft, review queue \\
\texttt{make run}     & full sweep, resumable \\
\texttt{make analyze} & five tables, three figures \\
\texttt{make paper}   & this PDF \\
\texttt{make verify}  & integrity checks \\
\texttt{make review}  & consistency report \\
\bottomrule
\end{tabular}
\caption{Build targets.}
\end{table}

\section{Item construction}
\label{sec:appendix-items}

Item identity is the position of a row in the two MetaTool Task2 files, written
\texttt{pair\_NNNN}. The 200 selected positions, the selection seed, the
MetaTool commit and the SHA-256 of each source file are recorded in
\texttt{data/manifest.json}. Selection is uniform without replacement over
positions; no item is dropped for being hard, ambiguous or awkward to
translate. The only permitted exclusion is translation drift, which is reported
as a rate with per-item reasons.

Korean prompts are produced by replacing both occurrences of the user query and
nothing else. The harness re-parses each produced prompt and asserts
byte-identity of the preamble, the tool list, the few-shot block and the
completion cue against the English original.

\section{Which names stay in English}
\label{sec:appendix-names}

Translating only the query leaves one decision that the byte-identity
assertion cannot make: what to do with a name inside the sentence being
translated. Both extremes are wrong, and they are wrong in opposite
directions.

Localizing everything breaks the item. A tool name transliterated into the
Korean alphabet (our discarded first draft did this to \texttt{Chatbot} and to
\texttt{Magnetis}) leaves a query asking about a tool that is not in the
candidate list, and no answer to it is correct. That failure scores as a model
error, in Korean only, on exactly the items whose query was most informative,
which biases in favor of the hypothesis.

Keeping everything in English breaks the condition. A Korean sentence with
\texttt{Japan} and \texttt{Paris} spliced into it is not what a Korean user
types, and the paper's premise is that the language of the query is what
changes the model's distribution. Worse, it changes it in a specific
direction: leaving English tokens in the query raises the probability the model
puts on English continuations, which are the continuations the enum grammar
allows. That lowers \texttt{masked\_mass} and shrinks the very quantity the
mechanism analysis is about, again in the direction that favors the
hypothesis, this time by suppressing the effect in Korean rather than
manufacturing it.

The line is therefore drawn at what the name is for. A name a tool would use
as a lookup key stays in English: tool names first, then brands, products,
people, teams and specific venues: \texttt{Marina Bay Sands},
\texttt{Arsenal}, \texttt{Suwon Station}, \texttt{Taylor Swift},
\texttt{New York Yankees}. Ordinary geography, languages and nationalities are
written in Korean, because they are ordinary vocabulary rather than keys: the
Korean words for New York City, Chicago, London, Paris and Russian. The
distinction cuts through individual cities. \texttt{New York City} is
localized where \texttt{New York Yankees} is not, because the line is about the
function of the string, not its type.

Thirty-seven of the 200 pairs were affected. Each was listed for the reviewer
with the name in question, and the set was accepted as translated; the release
records the affected items and the ruling.

Table~\ref{tab:koexamples} shows the rule applied. \texttt{Marina Bay Sands}
and \texttt{Magnetis} survive because a car-park lookup and a portfolio lookup
need those strings; \texttt{Paris} does not, because the Korean word is what a
Korean user would type and no tool matches on the English spelling.
\texttt{Sudoku} is both a tool name and an ordinary word, and stays in English
on the first ground. The last two rows carry no names at all and show the
speaker and sentence type being held fixed: a request stays a request in the
first person, and a question stays a question.

\begin{table*}[t]
\centering\small
\begin{tabular}{@{}l p{0.41\textwidth} p{0.41\textwidth}@{}}
\toprule
Item & English & Korean \\
\midrule
\texttt{pair\_0322} & I need a car park with available lots near Marina Bay
Sands. & \ko{저는 Marina Bay Sands 근처에 빈자리가 있는 주차장이 필요합니다.} \\
\addlinespace
\texttt{pair\_0249} & Can you please provide me with the current time in Paris,
France accurately? & \ko{프랑스 파리의 현재 시각을 정확하게 알려 주실 수
있나요?} \\
\addlinespace
\texttt{pair\_0732} & Can you update me on the current status of my portfolio
returns through Magnetis? & \ko{Magnetis를 통한 제 포트폴리오 수익의 현재
상태를 알려 주실 수 있나요?} \\
\addlinespace
\texttt{pair\_0538} & Hello, Sudoku pro! \ldots{} Can you provide me with a
Sudoku puzzle to tackle? & \ko{안녕하세요, Sudoku 고수님! \ldots{} 제가 풀어 볼
Sudoku 퍼즐을 하나 주실 수 있나요?} \\
\addlinespace
\texttt{pair\_0035} & Search for Twitter keywords associated with the
entertainment industry in Japan. & \ko{일본의 엔터테인먼트 산업과 관련된
Twitter 키워드를 검색해 주세요.} \\
\addlinespace
\texttt{pair\_0796} & Can you save my information so you can give me
personalized assistance? & \ko{저에게 개인 맞춤형 도움을 주실 수 있도록 제
정보를 저장해 주실 수 있나요?} \\
\addlinespace
\texttt{pair\_0093} & I'm clueless about what to get my dad for Father's Day.
Can you provide me with some exciting gift options? & \ko{아버지의 날에
아버지께 무엇을 드려야 할지 전혀 모르겠어요. 흥미로운 선물 선택지를 몇 가지
알려 주실 수 있나요?} \\
\bottomrule
\end{tabular}
\caption{Seven of the 200 items, English against Korean. Only the query is
translated; the ten candidate names, their descriptions, the instruction block
and the few-shot examples around it are the English original, byte for byte.}
\label{tab:koexamples}
\end{table*}

\section{Translation protocol}
\label{sec:appendix-translation}

Only the user query is translated. Tool names, tool descriptions, the
instruction block and the few-shot examples stay in English, because that is
the deployed condition the paper is about: an execution interface that does not
change when the user does.

The queries were drafted twice. The first draft was machine translation and was
discarded: review found errors that its screen could not see, including
transliterated tool and brand names, reversed speaker roles, questions rendered
as statements, mixed scripts, and one query translated into a third language
entirely. The second draft, the one used, was written to the instructions
released with it. Both are in the release so the difference can be inspected.

Four checks screen the result. Three are structural, on the Korean text itself:
whether a tool name still appears in English, whether any character is neither
Hangul nor ASCII that the source did not already contain, and whether a
question is still a question. The fourth is semantic: a round trip back to
English, compared by bag-of-words token $F_1$. It applies only to a machine
draft, since back-translating an authored sentence with its own author
reproduces the author's reading rather than testing it. The generating
procedure, the exact instructions, the thresholds and the per-check counts are
written into \texttt{data/manifest.json} by the translation step; none of them
is retyped here.

Flagging is not exclusion by machine. The whole set goes to review, ordered so
that the doubtful items come first: a lost tool name, then a structural
failure, then the judgment calls, then the rest. Review is by one native
Korean speaker over all 200 pairs, who may correct a translation, accept it, or
exclude it; the per-item review status is released. Evaluation of Korean is
refused by the harness until that pass is recorded, and the record is a file a
human creates.

A set of 50 natively authored Korean items is released as an empty schema
alongside the translations. It exists so that the obvious objection to a
translated benchmark, that it measures translationese rather than Korean, can
be answered by data rather than by argument. It is not filled by this work.

\section{Scoring rules}
\label{sec:appendix-scoring}

Matching is case-insensitive, restricted to the item's own ten candidates,
requires word boundaries, and resolves overlapping matches in favor of the
longer name. Word boundaries are what prevent \texttt{search} from matching
inside \texttt{ResearchFinder}, \texttt{calculator} inside
\texttt{Tax\_Calculator} and \texttt{form} inside \emph{information}; longest-
match resolution is what keeps \texttt{PDF\&URLTool} from also counting as
\texttt{URLTool}. Twenty hand-labeled cases fix this behavior in the test
suite, concentrated on the free condition where the model emits prose.

\section{Why the primary metric is paired}
\label{sec:appendix-metric}

The design originally registered an absolute difference of differences,
$\Delta_{\text{abstain}}(\mathrm{KO}) - \Delta_{\text{abstain}}(\mathrm{EN})$,
where $\Delta$ is constrained minus \textbf{contract-short} accuracy in
percentage points. It
was replaced before any confirmatory data existed, and the replacement is
recorded in \texttt{PREREG.md} with its reason and its commit.

The defect is arithmetic, not empirical. Abstention accuracy is bounded above,
so a language whose \textbf{contract-short} accuracy is higher has further to
fall, and
the difference of two such differences carries the gap between the two
baselines inside it. Whenever the languages start at different levels, the
absolute difference and the retained fraction can order them oppositely, and
both orderings are correct answers to different questions. A pilot on ten
hand-written Korean items showed exactly that pattern, which is what prompted
the amendment. Its figures are recorded in \texttt{PREREG.md} rather than here:
a hand-written fixture of that size is not evidence about a model, and does not
belong in a results narrative.

Pairing removes the leak rather than adjusting for it. Because both languages
are evaluated on the same items, one can condition on the items both answered
correctly unconstrained. Within that set both baselines are 100\% by
construction, and what remains is the question actually being asked: whether the
constraint breaks Korean abstentions that English abstentions survive. Items lost in both languages cancel, which is McNemar's observation:
they are evidence that the constraint is costly, not evidence about language.

The retired metric is still reported in Table~\ref{tab:paired}. Three measures
of one effect can disagree, and a reader shown only the one that was chosen
cannot tell whether the choice mattered.

The retention ratio is reported here rather than there. It divides constrained
accuracy by \textbf{contract-short} accuracy, which is informative while the
denominator is comfortably away from the floor and misleading once it is not. A
cell that gains under the constraint returns a ratio above one, and a cell that
starts near zero returns a large ratio from a small movement. Both appear in this
data. The absolute levels the ratio is built from are in Table~\ref{tab:main},
and it is those, not the ratio, that the paired statistic conditions on.

\begin{table*}[t]
\centering
\small
\begin{tabular}{llrrrrrrr}
\toprule
Model & Lang & Abst.\ free & Abst.\ short & Abst.\ constr. & Tool\ free & Tool\ short & Tool\ constr. & Ret. \\
\midrule
Qwen3-0.6B & English & 65.0 & 59.0 & 35.5 & 26.0 & 16.5 & 43.0 & 60.2 \\
 & Korean & 64.0 & 44.0 & 63.5 & 18.5 & 6.0 & 38.5 & 144.3 \\
Qwen3-1.7B & English & 77.0 & 74.5 & 50.5 & 33.8 & 34.0 & 45.0 & 67.8 \\
 & Korean & 73.0 & 61.5 & 48.5 & 30.9 & 30.5 & 41.0 & 78.9 \\
Qwen3-4B & English & 57.0 & 57.0 & 48.2 & 12.9 & 66.6 & 72.5 & 84.6 \\
 & Korean & 58.2 & 57.7 & 59.7 & 6.0 & 55.2 & 68.7 & 103.5 \\
\bottomrule
\end{tabular}
\vspace{2pt}
\begin{minipage}{\linewidth}\footnotesize Reported together because neither column alone distinguishes a judgment from a query-blind policy, and the two policies are mirror images. Always abstaining scores 100 on the abstention half and 0 on the other; always naming the first candidate scores 100 on the tool half and 0 on the abstention half, the gold tool being first in every row of this benchmark. Read as a pair, either one shows up as a collapse in the column it cannot reach, and neither appears here. A retention ratio above 100 means the cell was more accurate under the constraint than without it, which the ratio cannot express as a gain and the paired statistic can.\end{minipage}
\caption{Abstention accuracy (\emph{Abst.}) and tool-selection accuracy (\emph{Tool}), in percent, in each of the three conditions, with the retention ratio: constrained abstention accuracy over \textbf{contract-short}, which is the denominator throughout and is not the same baseline as \textbf{contract-free}.}
\label{tab:retention}
\end{table*}

\section{The chat template and the first token}
\label{sec:appendix-chat}

The prompts are issued as raw completions. MetaTool's \texttt{action\_prompt}
is a completion prompt: few-shot examples followed by \texttt{User query:
"\ldots" tool: }, where the next token is the answer. Issued instead through a
chat template, the model tends to open by restating the cue
(\texttt{tool: None}, then a justification), so the first generated token is
one no enum member begins with, and the grammar removes the whole first
position.

Table~\ref{tab:chat-endpoint} measures that on the three ladder models in both
languages, in the unconstrained condition, with \texttt{enable\_thinking} sent
false and every other setting held. The effect is in one direction everywhere:
the masked mass is higher under the template in all six cells and every
interval excludes zero. It is not, however, the same thing in every cell. The
two larger models saturate, their medians sitting at or above the level where
the variable has no variance left, and the smallest does not. The mechanism
is visible in the adjacent columns: the share of first tokens that begin a
candidate name falls under the template in every cell, and it falls furthest
exactly where the mass saturates.

Saturation is therefore a property of a model on a prompt rather than of the
endpoint alone, and the conclusion for this design is the same either way. The
mechanism regression in \S\ref{sec:analysis} needs a first-token distribution
that varies; on the chat path it does not vary enough to carry one, in the
models where it matters most. That is why the confirmatory design runs on
completions.

The choice has a cost, and it is worth stating plainly. Both generations of
these checkpoints are post-trained, and their model cards document chat usage
alone, and neither documents a text-completion format. Running them on raw
completions is off-label use, and the table shows what is given up by it. Under
the template the same models produce far fewer unreadable outputs and, for the
two larger ones, abstain more accurately than they do on the path this study
measures. The reported effects are therefore effects on a prompt format that a
deployment would not choose, and the mechanism variable is what buys that.

\begin{table*}[t]
\centering
\small
\begin{tabular}{llrrrrrrr}
\toprule
Model & Lang & chat & compl. & difference & chat & compl. & chat & compl. \\
\midrule
Qwen3-0.6B & English & 0.950 & 0.714 & \textbf{+0.236} \tiny{[+0.209, +0.269]} & 12.2 & 44.0 & 1.8 & 48.5 \\
Qwen3-0.6B & Korean & 0.852 & 0.801 & \textbf{+0.051} \tiny{[+0.017, +0.091]} & 18.5 & 30.8 & 1.5 & 65.2 \\
Qwen3-1.7B & English & 1.000 & 0.333 & \textbf{+0.667} \tiny{[+0.629, +0.707]} & 16.5 & 80.0 & 0.2 & 15.0 \\
Qwen3-1.7B & Korean & 1.000 & 0.661 & \textbf{+0.339} \tiny{[+0.315, +0.363]} & 3.0 & 56.2 & 0.0 & 31.2 \\
Qwen3-4B & English & 0.991 & 0.268 & \textbf{+0.723} \tiny{[+0.633, +0.771]} & 17.5 & 73.0 & 0.5 & 9.2 \\
Qwen3-4B & Korean & 1.000 & 0.504 & \textbf{+0.496} \tiny{[+0.395, +0.604]} & 3.2 & 55.0 & 0.0 & 18.8 \\
\bottomrule
\end{tabular}
\vspace{2pt}
\begin{minipage}{\linewidth}\footnotesize Columns three to five are \texttt{masked\_mass}; the difference is chat minus completions with a percentile interval resampling items, \textbf{bold} where it excludes zero. Not part of the confirmatory design: these records are held in a separate directory and \texttt{verify.py} refuses to admit two endpoints to one analysis.\end{minipage}
\caption{\textbf{Appendix.} The chat template against raw completions on the same items: median masked first-token mass, the share of first tokens that begin a candidate name (\%), and the share of outputs with no readable answer (\%). Same three models, both languages, the unconstrained condition, one seed, \texttt{enable\_thinking} false.}
\label{tab:chat-endpoint}
\end{table*}

\begin{table}[t]
\centering
\small
\begin{tabular}{lrrrrr}
\toprule
Lang & $n$ & Mean & Median & 10th pct. & $\geq 0.99$ \\
\midrule
English & 1200 & 0.831 & 0.986 & 0.282 & 46\% \\
Korean & 1200 & 0.893 & 0.997 & 0.592 & 57\% \\
\bottomrule
\end{tabular}
\vspace{2pt}
\begin{minipage}{\linewidth}\footnotesize Measured on the pilot model over the same paired items as the main experiment, in the unconstrained condition only. Compare Table~\ref{tab:mass}, which reports the same quantity on the completion endpoint the experiment uses.\end{minipage}
\caption{First-token probability mass the enum grammar would remove when the same prompts are issued through the chat template, unconstrained.}
\label{tab:chatsat}
\end{table}

The reason this belongs in the paper rather than in a commit message is that
the deployed configuration is the one that fails. Serving a schema-constrained
tool router through a chat endpoint is the ordinary thing to do, and it is
where the constraint and the template disagree hardest about what the first
token is for. It is the same collision the paper is about, met at the level of
the prompt format rather than the language.

\section{Exploratory analyses}
\label{sec:appendix-exploratory}

Everything in this section was added after the registered results were seen. It
is exploratory, it carries no confirmatory weight, and none of it revises the
verdicts in \S\ref{sec:results}. The registered analysis reports what it was
built to report; these tables describe the arrangement of the numbers it
produced, and are given without interpretation.

Three things are quantified. Table~\ref{tab:xpaths} reports how the
\texttt{masked\_mass} coefficient moves when a language term enters the model,
and decomposes the total effect of language on flipping into the part carried
by the mass and the part that is not. Table~\ref{tab:xregressions} takes
recovery, an abstention the constraint repairs rather than breaks, as a
dependent variable in its own right, and adds the item's
\textbf{contract-short} difficulty as a covariate to the flip model. Table~\ref{tab:xbands} reports the
flip rate of each language within bands of masked mass.

\begin{table*}[t]
\centering
\small
\setlength{\tabcolsep}{4pt}
\begin{tabular}{lrrrrrr}
\toprule
Model & mass alone & mass $|$ lang & $a$ & $a\times b$ indirect & $c'$ direct & $c$ total \\
\midrule
Qwen3-0.6B & -2.37 \tiny{[-5.55, +0.37]} & -1.79 \tiny{[-5.15, +1.21]} & \textbf{+0.05 \tiny{[+0.02, +0.08]}} & -0.09 \tiny{[-0.26, +0.07]} & \textbf{-0.89 \tiny{[-1.44, -0.41]}} & \textbf{-0.97 \tiny{[-1.55, -0.48]}} \\
Qwen3-1.7B & \textbf{+2.37 \tiny{[+1.27, +3.58]}} & \textbf{+5.04 \tiny{[+2.80, +8.12]}} & \textbf{+0.34 \tiny{[+0.31, +0.37]}} & \textbf{+1.70 \tiny{[+0.96, +2.78]}} & \textbf{-1.64 \tiny{[-2.88, -0.72]}} & +0.11 \tiny{[-0.23, +0.45]} \\
Qwen3-4B & \textbf{+8.39 \tiny{[+4.48, +14.41]}} & \textbf{+9.95 \tiny{[+5.67, +19.43]}} & +0.01 \tiny{[-0.01, +0.03]} & +0.10 \tiny{[-0.10, +0.33]} & \textbf{-2.09 \tiny{[-5.79, -1.00]}} & \textbf{-1.53 \tiny{[-3.39, -0.74]}} \\
\bottomrule
\end{tabular}
\vspace{2pt}
\begin{minipage}{\linewidth}\footnotesize Added after the registered analysis was run and carrying no confirmatory weight. The first two columns are the \texttt{masked\_mass} coefficient without and with a language term. $a$ is the effect of language on the mass, $a\times b$ the indirect path through it, $c'$ the direct path, $c$ the total effect with the mass omitted. Log-odds except $a$, which is on the mass scale; percentile bootstrap intervals resampling items and seeds. On a logistic outcome $a\times b$ and $c'$ do not sum to $c$.\end{minipage}
\caption{\textsc{Exploratory.} Suppression and the path decomposition of the language effect on flipping.}
\label{tab:xpaths}
\end{table*}

\begin{table*}[t]
\centering
\small
\begin{tabular}{llrrrr}
\toprule
Model & Model form & $n$ & \texttt{masked\_mass} & \textsc{ko} & difficulty \\
\midrule
Qwen3-0.6B & recover $\sim$ mass + lang & 970 / 410 & \textbf{+3.69 \tiny{[+1.59, +6.44]}} & \textbf{+1.70 \tiny{[+1.15, +2.41]}} & -- \\
Qwen3-0.6B & flip $\sim$ mass + lang + difficulty & 750 & -1.91 \tiny{[-5.33, +1.12]} & \textbf{-0.91 \tiny{[-1.47, -0.39]}} & +0.57 \tiny{[-0.85, +0.82]} \\
Qwen3-1.7B & recover $\sim$ mass + lang & 640 / 205 & +1.23 \tiny{[-0.13, +3.08]} & +0.30 \tiny{[-0.42, +1.10]} & -- \\
Qwen3-1.7B & flip $\sim$ mass + lang + difficulty & 1100 & \textbf{+5.09 \tiny{[+2.74, +7.88]}} & \textbf{-1.64 \tiny{[-2.95, -0.76]}} & +0.25 \tiny{[-0.74, +0.73]} \\
Qwen3-4B & recover $\sim$ mass + lang & 853 / 83 & \textbf{+2.09 \tiny{[+0.33, +4.25]}} & \textbf{+1.08 \tiny{[+0.01, +2.82]}} & -- \\
Qwen3-4B & flip $\sim$ mass + lang + difficulty & 960 & \textbf{+10.01 \tiny{[+5.79, +19.54]}} & \textbf{-1.95 \tiny{[-5.47, -0.97]}} & -0.60 \tiny{[-1.17, +1.15]} \\
\bottomrule
\end{tabular}
\vspace{2pt}
\begin{minipage}{\linewidth}\footnotesize Added after the registered analysis. A recovery is an abstention the constraint repairs: wrong without it, right with it. Its sample is the items at risk of repair - those the unconstrained condition got wrong - and $n$ gives at-risk and recovered counts. Difficulty is the item's mean unconstrained success rate across languages and seeds, centred. Coefficients are log-odds with bootstrap intervals; \textbf{bold} excludes zero.\end{minipage}
\caption{\textsc{Exploratory.} Recovery as its own outcome, and the flip model with the item's unconstrained difficulty as a covariate.}
\label{tab:xregressions}
\end{table*}

\begin{table*}[t]
\centering
\small
\begin{tabular}{llrrrrr}
\toprule
Model & Mass band & $n$ EN & flip \% EN & $n$ KO & flip \% KO & KO $-$ EN \\
\midrule
Qwen3-0.6B & 0.20--0.33 & 50 & 40.0 & 25 & 60.0 & +20.0 \\
Qwen3-0.6B & 0.33--0.44 & 35 & 42.9 & 40 & 25.0 & -17.9 \\
Qwen3-0.6B & 0.44--0.46 & 40 & 62.5 & 35 & 28.6 & -33.9 \\
Qwen3-0.6B & 0.46--0.52 & 40 & 100.0 & 35 & 14.3 & -85.7 \\
Qwen3-0.6B & 0.52--0.54 & 45 & 55.6 & 30 & 50.0 & -5.6 \\
Qwen3-0.6B & 0.54--0.57 & 50 & 60.0 & 25 & 0.0 & -60.0 \\
Qwen3-0.6B & 0.57--0.60 & 55 & 27.3 & 20 & 100.0 & +72.7 \\
Qwen3-0.6B & 0.60--0.63 & 20 & 50.0 & 55 & 18.2 & -31.8 \\
Qwen3-0.6B & 0.63--0.69 & 25 & 20.0 & 50 & 10.0 & -10.0 \\
Qwen3-0.6B & 0.69--0.77 & 15 & 33.3 & 60 & 25.0 & -8.3 \\
Qwen3-1.7B & 0.00--0.03 & 110 & 13.6 & 0 & -- & -- \\
Qwen3-1.7B & 0.03--0.07 & 110 & 22.7 & 0 & -- & -- \\
Qwen3-1.7B & 0.07--0.13 & 100 & 30.0 & 10 & 50.0 & +20.0 \\
Qwen3-1.7B & 0.13--0.20 & 80 & 43.8 & 30 & 16.7 & -27.1 \\
Qwen3-1.7B & 0.20--0.33 & 65 & 61.5 & 45 & 22.2 & -39.3 \\
Qwen3-1.7B & 0.33--0.40 & 30 & 100.0 & 80 & 50.0 & -50.0 \\
Qwen3-1.7B & 0.40--0.48 & 25 & 80.0 & 85 & 29.4 & -50.6 \\
Qwen3-1.7B & 0.48--0.59 & 25 & 80.0 & 85 & 41.2 & -38.8 \\
Qwen3-1.7B & 0.59--0.64 & 0 & -- & 110 & 45.5 & -- \\
Qwen3-1.7B & 0.64--0.79 & 5 & 100.0 & 105 & 61.9 & -38.1 \\
Qwen3-4B & 0.01--0.03 & 56 & 8.9 & 40 & 0.0 & -8.9 \\
Qwen3-4B & 0.03--0.05 & 45 & 0.0 & 50 & 0.0 & +0.0 \\
Qwen3-4B & 0.05--0.06 & 41 & 4.9 & 56 & 8.9 & +4.1 \\
Qwen3-4B & 0.06--0.08 & 50 & 0.0 & 46 & 0.0 & +0.0 \\
Qwen3-4B & 0.08--0.10 & 54 & 7.4 & 41 & 0.0 & -7.4 \\
Qwen3-4B & 0.10--0.13 & 48 & 2.1 & 49 & 0.0 & -2.1 \\
Qwen3-4B & 0.13--0.15 & 44 & 11.4 & 52 & 0.0 & -11.4 \\
Qwen3-4B & 0.15--0.20 & 57 & 21.1 & 39 & 12.8 & -8.2 \\
Qwen3-4B & 0.20--0.26 & 50 & 38.0 & 46 & 0.0 & -38.0 \\
Qwen3-4B & 0.26--0.63 & 35 & 71.4 & 61 & 13.1 & -58.3 \\
\bottomrule
\end{tabular}
\vspace{2pt}
\begin{minipage}{\linewidth}\footnotesize Added after the registered analysis. Bands are deciles of the masked mass pooled over both languages, so a band holds the same range of mass for each; the two distributions overlap only in part, which is why some bands are empty on one side. Rates are over the conditional set of Table~\ref{tab:paired}. No interval is given: several bands hold few observations and the table is descriptive.\end{minipage}
\caption{\textsc{Exploratory.} Flip rate within bands of masked first-token mass, by language.}
\label{tab:xbands}
\end{table*}

\subsection{The effect against the level it is taken from}
\label{sec:appendix-cells}

\begin{figure}[t]
  \centering
  \resultfigure{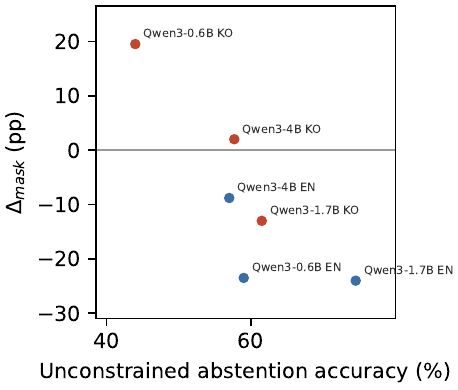}{\columnwidth}
  \caption{\textbf{Exploratory.} The masking effect against the
  \textbf{contract-short} abstention accuracy it is taken from, one point per
  (model, language) cell.
  Bars are the spread across seeds. The dotted line, when the fit is reported,
  marks where the fitted line crosses zero.}
  \label{fig:x-cells}
\end{figure}

Every quantity in this paper is a difference, and a difference is bounded by
the level it is taken from. A cell that abstains correctly less than half the
time without the constraint has little to lose to it; one that abstains almost
always has little to gain. Table~\ref{tab:x-cells} regresses the masking
effect on that level, one point per (model, language) cell, and
Figure~\ref{fig:x-cells} plots it. A negative slope would say the constraint
pulls toward a level rather than costing a fixed amount, and the point where
the fitted line crosses zero estimates that level.

The unit is the cell and the seeds are the variance within it, so the interval
resamples cells and then a seed inside each. The fit is withheld unless there
are five cells \emph{and} five models per predictor: the two cells of one model
are one model measured twice, and a line through a handful of points that share
a family is a picture rather than a finding.

\begin{table*}[t]
\centering
\small
\begin{tabular}{lrrrrrr}
\toprule
Terms & cells & Intercept (pp) & Baseline & \texttt{masked\_mass} & $R^2$ & Zero crossing (\%) \\
\midrule
baseline & 6 & -- & -- & -- & -- & -- \\
baseline + mass & 6 & -- & -- & -- & -- & -- \\
\bottomrule
\end{tabular}
\vspace{2pt}
\begin{minipage}{\linewidth}\footnotesize Intervals resample cells with replacement and then one seed within each cell; \textbf{bold} excludes zero. The zero crossing is $-a/b$, the unconstrained accuracy at which the constraint neither helps nor hurts. Not registered, and the cells share a benchmark and in places a family, so the intervals understate the dependence between them.\end{minipage}
\caption{\textbf{Exploratory.} The masking effect regressed on the unconstrained abstention accuracy it is taken from, one point per (model, language) cell. Every coefficient is withheld: this fit is over the three registered models, and the guard asks for five cells and five models per predictor before a coefficient is reported, so the row is shown empty rather than filled from six cells sharing one family.}
\label{tab:x-cells}
\end{table*}

\paragraph{Which models these cells are, and why.}
The decomposition below is fitted on \resultnum{census/models} models in two
languages, one cell each, against the three models the preregistration names.
Table~\ref{tab:x-cells} above is the same regression restricted to those three,
which is why its coefficients are withheld. The larger set is every model in the
released registry that produced a complete pair of conditions for the primary
contrast, which is all of them. The expansion was chosen for license, size and
family coverage before any of it was scored, and no model was added or dropped
on the strength of its result. The two scopes are kept apart throughout. The
confirmatory analysis in \S\ref{sec:analysis} reads the three registered models
only, its provenance records that scope, and its numbers do not move when the
other thirteen are present. Nothing in this appendix carries confirmatory
weight.

Whether the verdict depends on which models are in it is a fair question, and
Table~\ref{tab:x_decomp_robust} answers it by refitting on subsets that drop
whole tiers. The interval covers zero in every subset, so it does not.

\begin{table}[t]
\centering
\small
\begin{tabular}{lrr}
\toprule
Cells fitted & $n$ & \texttt{D0} $-$ \texttt{nonD0} [95\% CI] \\
\midrule
all cells & 32 & +0.083 [-0.186, +0.300] \\
without tier B & 26 & +0.063 [-0.418, +0.396] \\
without tier E & 26 & +0.087 [-0.203, +0.353] \\
without tier B and E & 20 & +0.181 [-0.293, +0.684] \\
\bottomrule
\end{tabular}
\caption{The difference between the two coefficients of the decomposition, refitted on subsets of the model set. Every subset is reported rather than a chosen one, and the interval covers zero in all of them, so the verdict does not depend on which tiers are included. Exploratory.}
\label{tab:x_decomp_robust}
\end{table}

\paragraph{Which half of the baseline's error the constraint repairs.}
That the level predicts the effect invites a reading with nothing to do with
format: a cell that got more wrong has more to gain, whatever the reason it got
things wrong. Table~\ref{tab:x_decomposition} tests that reading by splitting
the baseline's error into the share of items with no readable answer at all and
the share with a readable but incorrect one. The two sum to one minus the
baseline by construction, so a model containing both is the baseline model
rearranged unless the two carry different weights.

They do not. Both halves predict the contrast, their coefficients cannot be
told apart, and the pair explains no more than the baseline alone. The
constraint's benefit tracks how much the unconstrained condition got wrong and
not what kind of wrong it was, which is what the headroom reading predicts and
is not what a format-specific account predicts. We report it because it is the
result that bears hardest on the interpretation in \S\ref{sec:analysis}, and
because the transition counts in \S\ref{sec:results}, which are item-level
rather than cell-level, are the evidence that survives it.

\begin{table*}[t]
\centering
\small
\begin{tabular}{lllrr}
\toprule
Model & Term & Coefficient [95\% CI] & $R^2$ & AIC \\
\midrule
baseline & baseline & \textbf{-0.581 [-0.788, -0.395]} & 0.523 & -108.6 \\
D0 & D0 & \textbf{+0.454 [+0.215, +0.662]} & 0.222 & -93.0 \\
nonD0 & nonD0 & \textbf{+0.357 [+0.080, +0.632]} & 0.146 & -90.0 \\
D0 + nonD0 & D0 & \textbf{+0.624 [+0.378, +0.857]} & 0.528 & -106.9 \\
 & nonD0 & \textbf{+0.542 [+0.298, +0.780]} &  &  \\
\bottomrule
\end{tabular}
\caption{The unconstrained baseline's error split into the share of items with no readable answer (\texttt{D0}) and the share with a readable but wrong one (\texttt{nonD0}); the two sum to one minus the baseline by construction. Bold marks an interval that excludes zero. Exploratory.}
\label{tab:x_decomposition}
\end{table*}

The generators are \path{src/exploratory.py} and
\path{src/baseline_decomposition.py}. Their values are computed from the same
records as everything else, and \texttt{make verify} regenerates these tables
along with the registered ones.

\section{Request outcomes and the execution environment}
\label{sec:appendix-bookkeeping}

Table~\ref{tab:failures} reports request outcomes. Failed requests are excluded
from accuracy rather than scored as errors, so that an infrastructure failure is
not charged to the model; the counts are given so that the exclusion is visible.
An output carrying a reasoning block is not an infrastructure failure and is
scored wrong rather than dropped, for the reason given in \S\ref{sec:setup}.

\begin{table*}[t]
\centering
\small
\begin{tabular}{lllr}
\toprule
Model & Lang & Condition & ok \\
\midrule
Qwen3-0.6B & English & Constrained & 2000 \tiny{(100.0\%)} \\
Qwen3-0.6B & English & Free & 2000 \tiny{(100.0\%)} \\
Qwen3-0.6B & English & Short & 2000 \tiny{(100.0\%)} \\
Qwen3-0.6B & Korean & Constrained & 2000 \tiny{(100.0\%)} \\
Qwen3-0.6B & Korean & Free & 2000 \tiny{(100.0\%)} \\
Qwen3-0.6B & Korean & Short & 2000 \tiny{(100.0\%)} \\
Qwen3-1.7B & English & Constrained & 2000 \tiny{(100.0\%)} \\
Qwen3-1.7B & English & Free & 2000 \tiny{(100.0\%)} \\
Qwen3-1.7B & English & Short & 2000 \tiny{(100.0\%)} \\
Qwen3-1.7B & Korean & Constrained & 2000 \tiny{(100.0\%)} \\
Qwen3-1.7B & Korean & Free & 2000 \tiny{(100.0\%)} \\
Qwen3-1.7B & Korean & Short & 2000 \tiny{(100.0\%)} \\
Qwen3-4B & English & Constrained & 2000 \tiny{(100.0\%)} \\
Qwen3-4B & English & Free & 2000 \tiny{(100.0\%)} \\
Qwen3-4B & English & Short & 2000 \tiny{(100.0\%)} \\
Qwen3-4B & Korean & Constrained & 2000 \tiny{(100.0\%)} \\
Qwen3-4B & Korean & Free & 2000 \tiny{(100.0\%)} \\
Qwen3-4B & Korean & Short & 2000 \tiny{(100.0\%)} \\
\bottomrule
\end{tabular}
\caption{Request outcomes by cell. Non-\texttt{ok} requests are excluded from accuracy rather than counted as errors, so that infrastructure failure is not attributed to the model.}
\label{tab:failures}
\end{table*}

Table~\ref{tab:env} records the execution environment. Results from many hosts
are aggregated into single cells, so a single row here is a precondition for the
aggregation being sound: more than one column would mean environment differences
are confounded with seed variance.

\begin{table*}[t]
\centering
\small
\begin{tabular}{lp{0.62\textwidth}}
\toprule
Field & Value \\
\midrule
vllm & 0.27.1 \\
torch & 2.13.0 \\
cuda & 13.0 \\
gpu & NVIDIA L4 \\
driver & 595.71.05 \\
api\_path & structured\_outputs \\
endpoint & completions \\
model\_revision & -- \\
server\_env & VLLM\_USE\_FLASHINFER\_SAMPLER=0 \\
records & 36000 \\
\bottomrule
\end{tabular}
\vspace{2pt}
\begin{minipage}{\linewidth}\footnotesize A single column means every number in this paper came from one software and hardware configuration. More than one means environment differences are confounded with seed variance and the aggregation is not sound.\end{minipage}
\caption{Execution environment of the recorded results.}
\label{tab:env}
\end{table*}

\section{Schema-translation control}
\label{sec:appendix-schema}

An additional condition in which tool \emph{descriptions} are also translated
into Korean, with tool names left as English identifiers, is produced by the
release but is not part of the confirmatory design. It separates the effect of
a non-English user query from the effect of a non-English schema.

\emph{Not yet run.}

\section{Relation to the nearest antecedents}
\label{sec:appendix-antecedents}

Three of the works \S\ref{sec:related} names are near neighbors, and what
separates them from this paper is what stops the contributions reading as
rediscovery. \S\ref{sec:intro} and \S\ref{sec:related} state each
separation in a sentence; this appendix works all three through.

\subsection{Structure snowballing}

The second contribution has an antecedent that reaches the same shape of
conclusion. \citet{zhou2026snowballing} enforces a reflection schema on a model
asked to critique its own reasoning, and finds that it does not improve
self-correction: the agent achieves near-perfect surface conformance while the
deeper semantic errors go undetected. He names this structure snowballing and
attributes it to the cognitive load of satisfying strict format rules, which
pushes the model into a formatting trap. Surface form is bought and judgment
is not, which is what this paper reports as well.

\begin{table}[t]
  \centering\small
  \begin{tabular}{@{}p{0.19\columnwidth}p{0.34\columnwidth}p{0.34\columnwidth}@{}}
    \toprule
     & \citet{zhou2026snowballing} & This paper \\
    \midrule
    Task & Self-critique, open-ended & Tool abstention, closed choice \\
    Models & One, 8B & \resultnum{census/complete-grid-cap} registered,\\ \resultnum{census/models} in total, from 0.6B \\
    Languages & English & English and Korean \\
    Constraint & Applied whole & Stop and enum separated \\
    Evidence & Failure mode named & Item-level transition counts \\
    Mechanism & Cognitive load, hypothesized & Repair at the site of format collapse, measured \\
    \bottomrule
  \end{tabular}
  \caption{The nearest antecedent to the second contribution.}
  \label{tab:vs-snowballing}
\end{table}

The last row of Table~\ref{tab:vs-snowballing} is the one that matters. His
cognitive-load account is a hypothesized mechanism for an observed failure
mode, offered as an explanation rather than measured. This paper puts a
decidable quantity in that place. Counting what each repaired output looked
like before the constraint was applied turns the question into arithmetic. Of
\resultnum{repair/n-ft} repaired abstentions, \resultnum{repair/n-ft-judgement}
were judgments the scorer refused, which rules out the reading that the judgment
was present and the format was blocking it. In the cell with the most repair,
\resultnum{repair/cell-n-ft-unreadable} of \resultnum{repair/cell-n-ft} had no
readable output there to begin with. The contribution is not that we reach his conclusion again; it
is that the mechanism he leaves as a hypothesis is adjudicated at the level of
items.

\subsection{Two constraint taxes with the same name}

\citet{li2026suppression} is the sharpest result in the cluster, and the cause
is mechanical. The schema compiles to an automaton in which the angle bracket
that opens a Qwen-style \texttt{<tool\_call>} tag is forbidden in every state,
so the tag is unreachable at every position and the behavior it would have
introduced cannot occur. Neither supervised fine-tuning nor reinforcement
learning recovers it, which is what a mask downstream of the weights predicts.

That is a different measurement from ours, and the difference is the whole
reason both are worth having. Their constraint and the behavior it suppresses
lie on separate axes: the schema governs the shape of a reply while the tool
call is a thing the model does, and the mask happens to make the tokens for it
unreachable. Ours is an enum over the answer set itself, and the abstention
option is inside it by construction, because a grammar that omits the
abstention option makes abstention impossible and the measurement vacuous
(\S\ref{sec:setup}). Reporting that a model does not produce an answer whose
tokens are masked is a statement about the decoder; reporting whether a model
still declines when declining remains available is a statement about the model.

The two results also behave differently across models. Their outcome is uniform,
as a mechanical consequence should be.
Ours is not uniform once the intervention is taken apart: the enum component
$\Delta_{\text{mask}}$ changes sign from cell to cell, which a mechanical
consequence would not. Measured whole, against free generation, our effect on
abstention does keep one sign; it is the decomposition that finds a term
pointing the other way, and a mask that makes the answer unreachable leaves
nothing to decompose.

\begin{table}[t]
  \centering\small
  \begin{tabular}{@{}p{0.20\columnwidth}p{0.34\columnwidth}p{0.34\columnwidth}@{}}
    \toprule
     & \citet{li2026suppression} & This paper \\
    \midrule
    Constraint & JSON Schema, on an axis separate from the tool call
               & Enum over the answer set itself \\
    Correct token & Unreachable under the mask
               & Present in the enum by construction \\
    Result & Uniform across models
               & $\Delta_{\text{mask}}$ changes sign between cells \\
    Measures & A property of the decoder
               & A property of the model's judgment \\
    \bottomrule
  \end{tabular}
  \caption{Two constraint taxes with the same name and different subjects.}
  \label{tab:vs-suppression}
\end{table}

Table~\ref{tab:vs-suppression} sets the two designs side by side. Their
taxonomy of suppressed behaviors needs one comparison in addition, because one
of its categories looks like ours. Their \textsc{TS-C}, intent without action,
covers a reply that states a tool is needed without emitting the call. Our
\texttt{D2} covers an output naming two or more enum members, which includes
the case of naming a tool and abstaining in one breath. The two are close
enough to be confused and are not the same: \textsc{TS-C} is a reply that could
not act because the action was masked away, and \texttt{D2} is a reply that
could have acted and did not resolve to one answer. That we find no
\texttt{D2} containing the abstention token among the outputs the constraint
turned from wrong to right (\S\ref{sec:results}) is a statement about ours that
does not transfer to theirs.

\citet{ray2026constrainttax} is the closer of the two in scope, and is where
the phrase constraint tax comes from. His calendar tool-call analogue is the
nearest published setting to ours, and it makes the point that the residual
errors are semantic rather than structural: a prompt-only schema and an
enforced one reach the same full validity at very different executable
accuracy. Three things separate it from this paper. He does not measure
abstention, which is the behavior an enum over tool names is most likely to
remove. He has no second language, so the question of whether the cost is
language-dependent cannot arise. And he does not decompose the constraint,
so a tax is a tax everywhere he looks; separating the stop token from the enum
is what turns a single negative number into two components with opposite signs.
His reporting recommendation, that schema validity and answer accuracy be given
separately rather than combined, is one we follow throughout.

\section{The shipped candidate ordering}
\label{sec:appendix-ordering}

\S\ref{sec:results} states that the ordering cancels out of every reported
difference. This is the argument for it.

Ruling out a model that always abstains leaves the opposite policy to rule out.
The gold tool is the first candidate in every row of this benchmark
(\S\ref{sec:setup}), so a decoder that pulled the model toward the head of the
list would raise tool-selection accuracy mechanically, with no improvement in
judgment. The subset that would settle this directly, items whose gold is not
first, does not exist: it is \resultnum{position/gold-not-first-n} of
\resultnum{position/gold-n} items, which is a property of the release and is a
limit on what this benchmark can be asked (Limitations).

What can be measured is the size of the pull, on the half where naming the
first candidate is always wrong. On abstention items the share of outputs
naming the first candidate moves by \resultnum{position/abstain-shift-mean}
points on average between the unconstrained and constrained conditions, at most
\resultnum{position/abstain-shift-max} in any cell. And on tool-needed items
the constrained condition names the first candidate no more often than leaving
the decoder alone entirely does, in
\resultnum{position/tool-not-above-free} of \resultnum{dmask/abstain/n-cells}
cells. The stop token pushed the models off the first candidate, and the enum
returns them to roughly where free generation already had them.

Those two bound the pull, and they bound it at something small. The test they
stand in for, moving the gold off the first position and re-measuring, cannot
be run on our data, because no item has its gold anywhere else. It can
be run on permutations of the data, and a companion note runs it
\citep{positionbias2026}. Across ten positions, four models and English, no
model answers by naming the first candidate, and accuracy is lower when the gold
is first than when it is elsewhere.

That reverses the direction of the concern. The shipped ordering is not a
tailwind for tool selection on these models; it is a headwind, and the
tool-selection accuracies reported here should be read as a lower bound rather
than an inflated one. It does not follow that the contrasts need adjusting. The
reason is the one that makes this design work at all: the ordering is a property
of the item and is byte-identical across the three conditions. Whatever it does
to a level it does equally to every condition, so it cancels out of
$\Delta_{\text{mask}}$, $\Delta_{\text{length}}$ and $\Delta_{\text{total}}$.
The confound reaches absolute levels and not differences. The note's evidence is also four models in one language on
permutations of the benchmark rather than the benchmark, so it bounds the
concern rather than closing it.

\section{Both item types together}
\label{sec:appendix-pooled}

\S\ref{sec:results-pooled} reports the pooled column and why the abstention
column leads. This appendix prints the table and carries the rest of that
comparison.

\begin{table*}[t]
\centering
\small
\begin{tabular}{lrrr}
\toprule
Cell & Abstention & Tool selection & Pooled \\
\midrule
Qwen3-0.6B EN & -29.5 & +17.0 & -6.2 \tiny{[-9.4, -3.2]} \\
Qwen3-0.6B KO & -0.5 & +20.0 & +9.8 \tiny{[+6.7, +12.8]} \\
Qwen3-1.7B EN & -26.5 & +11.2 & -7.7 \tiny{[-10.7, -4.6]} \\
Qwen3-1.7B KO & -24.5 & +10.1 & -7.2 \tiny{[-10.2, -4.0]} \\
Qwen3-4B EN & -8.8 & +59.6 & +25.4 \tiny{[+22.4, +28.3]} \\
Qwen3-4B KO & +1.5 & +62.7 & +32.1 \tiny{[+29.1, +35.0]} \\
\midrule
Mean & -14.7 & +30.1 & +7.7 \\
\bottomrule
\end{tabular}
\vspace{2pt}
\begin{minipage}{\linewidth}\footnotesize The pooled column does not have the same sign as the abstention column, and the paper's headline is the abstention column. \S\ref{sec:results} says why that is the one to lead with, and the reason is not that it is the more favorable of the two.\end{minipage}
\caption{$\Delta_{\text{total}}$, constrained minus contract-free, in accuracy points: on abstention items, on tool-needed items, and pooled over both. The two item types are equinumerous by construction, so the pooled column is an unweighted mean over 400 items and needs no weighting choice.}
\label{tab:pooled}
\end{table*}

Table~\ref{tab:pooled} pools the two item types. They are equinumerous by
construction, so the pooled column is an unweighted mean over 400 items and
there is no weighting to argue about. \textbf{It does not have the same sign as
the abstention column.} The mean over cells is \resultnum{pooled/mean} points,
against \resultnum{dtotal/abstain/min} at worst on abstention alone. It is
negative in \resultnum{pooled/n-negative} of
\resultnum{dmask/abstain/n-cells} cells, where the abstention column is negative
in \resultnum{abstain/n-negative}. Every headline in this paper is the
abstention column, and the pooled column is here so that the difference is
stated by us rather than discovered.

The reason to lead with abstention is not that it is the more favorable half,
and it is not a claim that the other half is uninterpretable. It is that
abstention accuracy is what the preregistration names as the primary measure.
Moving a headline to a different metric after seeing which one reads better is
the behavior preregistration exists to prevent, and the fact that the pooled
number happens to be kinder to the intervention makes moving to it worse rather
than better. Both are reported here; the registered one leads.

Two further considerations point the same way. The tool column's absolute level
is depressed by the shipped candidate ordering rather than inflated by it, so
it is a poor choice of headline for a paper that reports levels alongside
differences (\S\ref{sec:results}). And the evidence separating selection from
position is four models in one language on permuted copies of the benchmark
\citep{positionbias2026}, which is narrower than the confirmatory scope it
would have to support.

\paragraph{Against the free baseline the two columns move apart.}
The pattern in Table~\ref{tab:pooled} is tool selection up in every cell and
abstention down in \resultnum{abstain/n-negative} of
\resultnum{dmask/abstain/n-cells}. That is the signature of a decision threshold
moving toward naming a tool, and it deserves to be named rather than left for a
reader to infer. It is not the same phenomenon as the repair above, and the two
belong to different components. Measured from the stop token, the grammar gives
back what that stop token destroyed, and both columns rise. Measured from free
generation, the grammar also shifts where the model draws the line between
acting and declining, and that shift pays on the tool half and costs on the
abstention half.

The shift is real but it is not uniform, and the mechanism differs by model. On
abstention items the share of outputs naming a tool rises by seventeen to
twenty-four points from free to constrained for the 0.6B and 1.7B models, which
is a threshold moving toward action. For the 4B model that share falls instead,
so its tool-selection gain comes from somewhere else: its free-generation
answers name several tools at once, and the grammar forces exactly one. Calling
the whole pattern a threshold shift would be tidier than the data.

This is the decomposition argument rather than an exception to it. One
intervention that a two-condition design reports as a single number is doing at
least two separable things with opposite signs on the abstention column:
repairing format damage, which helps, and moving the decision boundary toward
naming a tool, which hurts. Reporting them apart is the point of the third
condition.

\section{Mass distributions and the size trend}
\label{sec:appendix-massfigs}

The second registered analysis and two secondary views of it. The regression
asks whether the mass predicts an item's flip and whether adding a language
term explains anything the mass has not; the coverage table is what makes the
mass a usable lower bound.

\begin{table*}[t]
\centering
\small
\begin{tabular}{llrrrr}
\toprule
Model & Terms & $n$ & flips & \texttt{masked\_mass} & \textsc{ko} \\
\midrule
Qwen3-0.6B & mass & 750 & 295 & -2.37 \tiny{[-5.55, +0.37]} \tiny{$p=0.089$} & -- \\
Qwen3-0.6B & mass + lang & 750 & 295 & -1.79 \tiny{[-5.15, +1.21]} \tiny{$p=0.246$} & \textbf{-0.89 \tiny{[-1.48, -0.40]} \tiny{$p<0.001$}} \\
Qwen3-1.7B & mass & 1100 & 455 & \textbf{+2.37 \tiny{[+1.27, +3.58]} \tiny{$p<0.001$}} & -- \\
Qwen3-1.7B & mass + lang & 1100 & 455 & \textbf{+5.04 \tiny{[+2.80, +8.12]} \tiny{$p<0.001$}} & \textbf{-1.64 \tiny{[-2.92, -0.74]} \tiny{$p<0.001$}} \\
Qwen3-4B & mass & 960 & 91 & \textbf{+8.39 \tiny{[+4.48, +14.41]} \tiny{$p<0.001$}} & -- \\
Qwen3-4B & mass + lang & 960 & 91 & \textbf{+9.95 \tiny{[+5.67, +19.43]} \tiny{$p<0.001$}} & \textbf{-2.09 \tiny{[-5.52, -1.02]} \tiny{$p<0.001$}} \\
\bottomrule
\end{tabular}
\vspace{2pt}
\begin{minipage}{\linewidth}\footnotesize Coefficients are log-odds with percentile bootstrap intervals resampling items and seeds; \textbf{bold} excludes zero. Fitted on the conditional set of Table~\ref{tab:paired}, one observation per item, language and seed. A language term whose interval covers zero says the language acts through the mass and not beside it; one that excludes zero says the mass does not account for it.\end{minipage}
\caption{Logistic regression of \emph{flip} --- an abstention held without the constraint and lost with it --- on the first-token probability mass the grammar removes, with and without a language term.}
\label{tab:mechanism}
\end{table*}

\begin{table*}[t]
\centering
\small
\begin{tabular}{llrrrrr}
\toprule
Model & Lang & Mean mass & Median mass & $n$ & Slope & $R^2$ \\
\midrule
Qwen3-0.6B & English & 0.685 & 0.695 & 200 & +1.069 & 0.059 \\
Qwen3-0.6B & Korean & 0.754 & 0.771 & 200 & +2.272 & 0.163 \\
Qwen3-1.7B & English & 0.386 & 0.382 & 200 & +0.439 & 0.027 \\
Qwen3-1.7B & Korean & 0.634 & 0.661 & 200 & +0.889 & 0.057 \\
Qwen3-4B & English & 0.375 & 0.365 & 200 & +0.121 & 0.006 \\
Qwen3-4B & Korean & 0.516 & 0.520 & 200 & +0.283 & 0.030 \\
\bottomrule
\end{tabular}
\vspace{2pt}
\begin{minipage}{\linewidth}\footnotesize \texttt{masked\_mass} is the first-token probability mass that the enum grammar would forbid, measured in the unconstrained condition. Because it is summed over a truncated top-$k$ head of the distribution it is a lower bound; the top-$k$ head covered 0.982 of the distribution on average. Slope and $R^2$ are for the per-item regression of $\Delta_{\text{abstain}}$ on \texttt{masked\_mass}.\end{minipage}
\caption{Masked first-token probability mass and its relation to the abstention penalty.}
\label{tab:mass}
\end{table*}

\begin{figure*}[t]
  \centering
  \resultfigure{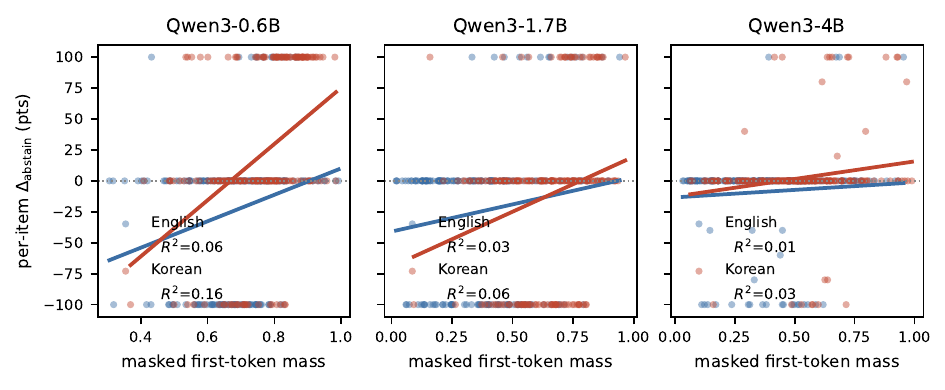}{\textwidth}
  \caption{Per-item abstention penalty against the first-token probability mass
  the enum grammar removes, by language, with least-squares fits.}
  \label{fig:mechanism}
\end{figure*}

\begin{figure*}[t]
  \centering
  \resultfigure{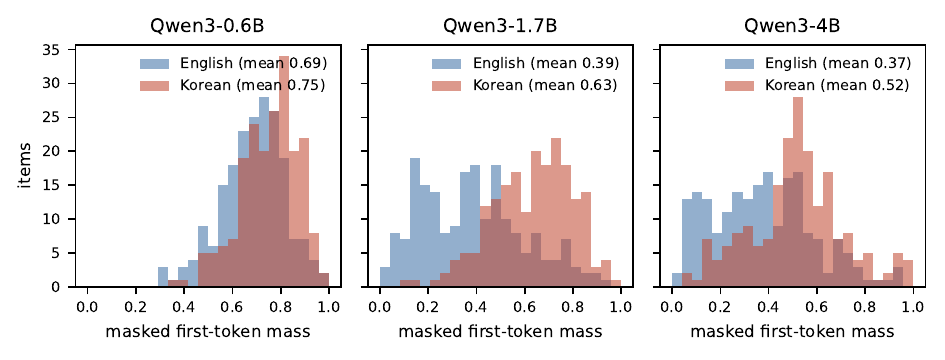}{\textwidth}
  \caption{Distribution of masked first-token probability mass, English against
  Korean.}
  \label{fig:mass}
\end{figure*}

\begin{figure}[t]
  \centering
  \resultfigure{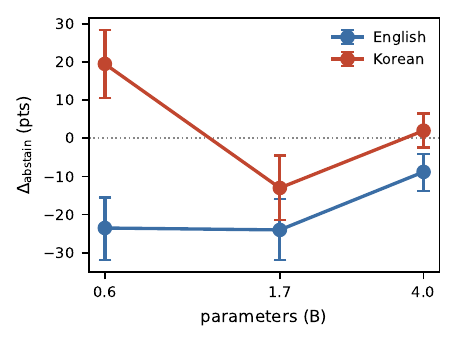}{\columnwidth}
  \caption{Abstention penalty against model size.}
  \label{fig:size}
\end{figure}

Figure~\ref{fig:mechanism} plots the per-item penalty against the mass,
Figure~\ref{fig:mass} shows the mass distributions directly, and
Figure~\ref{fig:size} plots the penalty against parameter count.

Registered as a secondary observation: whether the penalty grows as models
shrink. Figure~\ref{fig:size} plots $\Delta_{\text{abstain}}$, the
registration's name for $\Delta_{\text{mask}}$ on abstention items (constrained
minus \textbf{contract-short}), against parameter count for each language.

\section{Properties of the released benchmark}
\label{sec:appendix-release}

\S\ref{sec:setup} names the four properties in a sentence. Each is stated
here with what it does and does not threaten.

Four further properties bear on interpretation, and we did not find them
stated in the benchmark's documentation. First, the ten candidates offered to a
paired item are \emph{disjoint} between the two files: mean overlap is zero
across all 995 pairs, so the pair controls the query but not the distractor
set. Second, the few-shot block differs systematically by item type: every
Subtask3 prompt contains exactly two examples labeled \texttt{None}, and every
Subtask1 prompt contains none, so the abstention items are primed toward
abstention in a way the tool-needed items are not. Third, the user query
appears twice in each prompt, once in the query block and once in the trailing
completion cue. The Limitations section states what each of these does and does
not threaten.

Fourth, the candidate list is ordered rather than shuffled. In all 995 Subtask1
rows the gold tool is the first of the ten candidates the prompt offers, and in
our sample the abstention token is the last member of the enum in all 200
items. A policy of naming the first candidate and nothing else therefore
answers every tool-needed item correctly and every abstention item incorrectly,
without reading the query at all. This does not disturb
$\Delta_{\text{mask}}$, which is a within-item difference between two
conditions that see the same ordering, and it cannot manufacture abstention
accuracy, since the abstention answer is not in the tool list at any position.
It does mean that tool-selection accuracy on this benchmark is not by itself
evidence of tool selection.

How far each model takes the ordering up varies far more than the models
themselves do. That is a property of the benchmark rather
than a result about constrained decoding, so it is quantified separately in a
companion note \citep{positionbias2026}; Limitations states what it costs the
tool-needed column here.

\section{What else was collected}
\label{sec:appendix-collected}

The full collection is larger and is not one thing, so we give it by what each
part supports rather than as a total.
\resultnum{census/confirmatory} requests are the three registered models above.
A further \resultnum{census/regression} are the abstention items of
\resultnum{census/models} models in the two conditions of the primary contrast,
which is what the exploratory cell regression reads
(\S\ref{sec:appendix-cells}). \resultnum{census/chat} measure the chat endpoint
against the completions endpoint (\S\ref{sec:appendix-chat}), and
\resultnum{census/native} are the natively authored Korean control
(\S\ref{sec:appendix-schema}).

\resultnum{census/unread} requests are read by no table in this paper. They are
the tool-needed items of the expansion models, and the reason they go unused is
the same reason the three-contrast argument lives on six cells:
\textbf{contract-free} was collected in full only for the three registered
models. Of \resultnum{census/models} models, \resultnum{census/complete-grid}
have all three conditions in both languages; the other thirteen are missing
\textbf{contract-free} entirely or have it for one seed. $\Delta_{\text{length}}$
and $\Delta_{\text{total}}$ are therefore not computable for them, and the
choice of which cells carry the decomposition is a constraint of the collection
rather than a selection made after seeing it.

Prompts are issued as raw completions, which is the form
the benchmark's prompt is written in; a chat template would put the model's
restatement of the completion cue in front of the answer and leave the
mechanism variable with no variance (\S\ref{sec:appendix-chat}). No chat
template means no reasoning block is reachable, and every record carries the
endpoint it was issued through so that the two can never be pooled.

\section{Scoring, in full}
\label{sec:appendix-scoringdetail}

\S\ref{sec:setup} states the rules and names the one departure. This is the
whole of it.

We adopt the single-match and ordering rules of \citet{lee2026patool}. For an
abstention item, a correct answer contains the abstention token and either no
candidate mention or a candidate mention that follows it. For a tool-needed
item, a correct answer mentions exactly the gold tool, ahead of any abstention
token. Our abstention rule is theirs exactly, and our \texttt{None}-first
counts are a direct consequence of it.

We depart from them in one place, and it is not a small one. Where an output
names two or more candidates, their procedure asks an LLM judge whether the
gold tool was the one selected; we score it wrong and count it, keeping the
whole pipeline deterministic. The two choices disagree on a real share of
outputs rather than on a handful, so we do not describe our scoring as theirs
without qualification. It is their rules for the cases a rule can settle, and a
deterministic loss for the cases they hand to a model. The direction of the
difference is that we are the stricter of the two, and the counts of
multi-candidate outputs are reported alongside the accuracies rather than
folded into them.

Applied literally, substring matching is unsafe on this candidate pool. Of the
199 tool names, 14 are substrings of another name, and several
(\texttt{search}, \texttt{form}, \texttt{Now}, \texttt{speak}) are ordinary
English words that occur naturally in the justification the free condition
elicits. We therefore require word-boundary matches and resolve overlapping
matches in favor of the longer name. Twenty hand-labeled cases pin this
behavior in the released test suite.

\end{document}